\pdfoutput=1
\documentclass[11pt]{article}
\usepackage{acl}
\usepackage{times}
\usepackage{latexsym}
\usepackage[T1]{fontenc}
\usepackage[utf8]{inputenc}
\usepackage{microtype}
\usepackage{booktabs}
\usepackage{graphicx}
\usepackage{amsmath}
\usepackage{enumitem}
\usepackage{changepage}
\usepackage{tabularx}
\usepackage{xcolor}
\usepackage{colortbl}
\usepackage{geometry}
\usepackage{changepage}
\usepackage{amsthm}
\usepackage{amsfonts}
\usepackage{amssymb}
\usepackage{thmtools}
\usepackage{mathtools}
\usepackage{xspace}
\usepackage{tcolorbox}
\usepackage{arydshln}
\usepackage{multirow}
\usepackage{hhline}
\usepackage{colortbl}
\usepackage{multicol}
\usepackage{floatrow}
\usepackage[capitalize,noabbrev]{cleveref}

\graphicspath{{figures/}{figures/dataset-examples/}{figures/model-performance/}{figures/data-analysis/}}

\setkeys{Gin}{width=\linewidth}

\newcommand{\fmn}[1]{\textbf{\texttt{#1}}}
\newcommand{\mn}[1]{\texttt{#1}}
\newcommand{\dn}[1]{\textit{#1}}  
\newcommand{\fdn}[1]{\textit{\textbf{#1}}}

\renewcommand{\texttt}[1]{%
  \begingroup
  \ttfamily
  \begingroup\lccode`~=`/\lowercase{\endgroup\def~}{/\discretionary{}{}{}}%
  \begingroup\lccode`~=`[\lowercase{\endgroup\def~}{[\discretionary{}{}{}}%
  \begingroup\lccode`~=`.\lowercase{\endgroup\def~}{.\discretionary{}{}{}}%
  \catcode`/=\active\catcode`[=\active\catcode`.=\active
  \scantokens{#1\noexpand}%
  \endgroup
}

\newcommand\blfootnote[1]{%
  \begingroup
  \renewcommand\thefootnote{}\footnote{#1}%
  \addtocounter{footnote}{-1}%
  \endgroup
}

\newcolumntype{C}[1]{>{\centering\arraybackslash}m{#1}}
\newcolumntype{L}[1]{>{\raggedright\arraybackslash}m{#1}}

\definecolor{gray}{rgb}{0.960,0.960,0.960}
\newcommand{\rg}[0]{\rowcolor{gray}}

\usepackage{setspace}
\makeatletter

\DeclareRobustCommand\onedot{\futurelet\@let@token\@onedot}
\def\@onedot{\ifx\@let@token.\else.\null\fi\xspace}

\makeatother

\newcommand{\wildqa}{\textsc{WildQA}}

\title{WildQA: In-the-Wild Video Question Answering}

\author{Santiago Castro\Thanks{~Equal contribution.} \quad Naihao Deng\footnotemark[1] \quad Pingxuan Huang\footnotemark[1] \quad Mihai Burzo \quad Rada Mihalcea \\
University of Michigan -- Ann Arbor, USA \\
\texttt{\{sacastro, dnaihao, pxuanh, mburzo, mihalcea\}@umich.edu}}

\begin{document}

\twocolumn[{%
\renewcommand\twocolumn[1][]{#1}%
\maketitle
\begin{center}
  \footnotesize
  \captionsetup{type=figure}
  \includegraphics{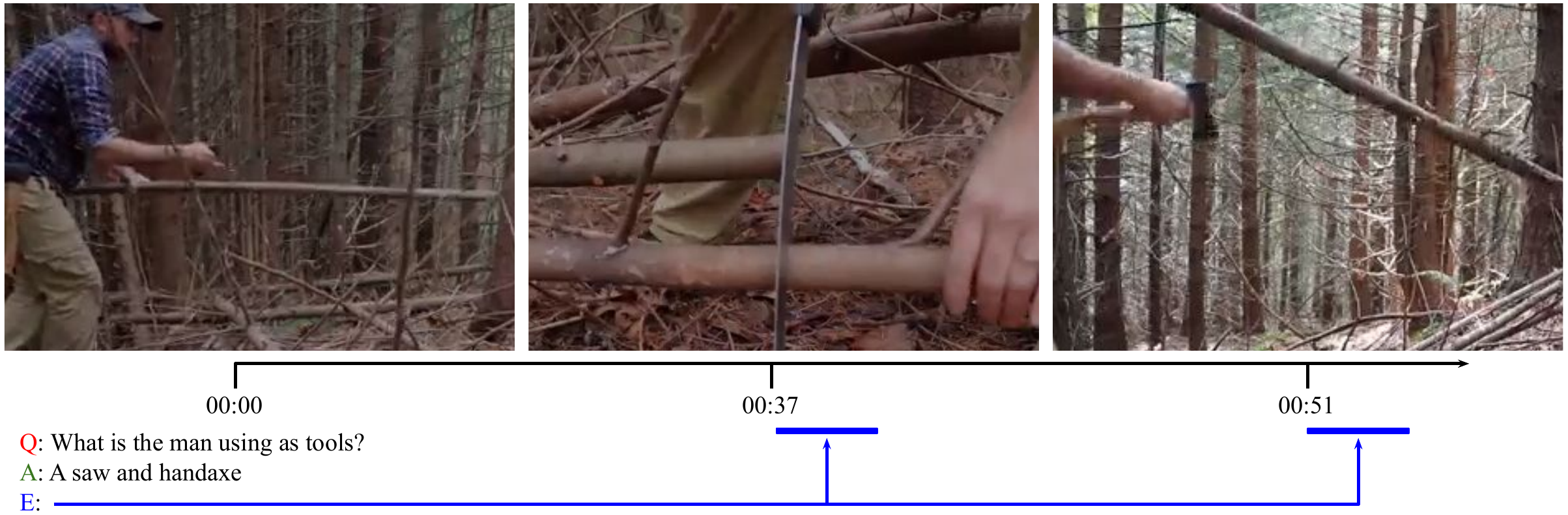}
    \caption{An example from our WildQA dataset, showing a question (Q), an answer (A), and evidence (E) that supports the answer. The corresponding part of the videos is provided as evidence for the question.}
\label{fig:evidence-finding-example}
\end{center}%
}]
\blfootnote{*: Equal contribution}

\begin{abstract}
Existing video understanding datasets mostly focus on human interactions, with little attention being paid to the "in the wild" settings, where the videos are recorded outdoors. We propose \textbf{\wildqa{}}, a video understanding dataset of videos recorded in outside settings. In addition to video question answering (Video QA), we also introduce the new task of  identifying visual support for a given question and answer (Video Evidence Selection). Through evaluations using a wide range of baseline models, we show that \wildqa{} poses new challenges to the vision and language research communities. The dataset is available at \url{https://lit.eecs.umich.edu/wildqa/}.

\end{abstract}

\section{Introduction}

\begin{figure*}
\centering
\includegraphics[width=0.8\textwidth]{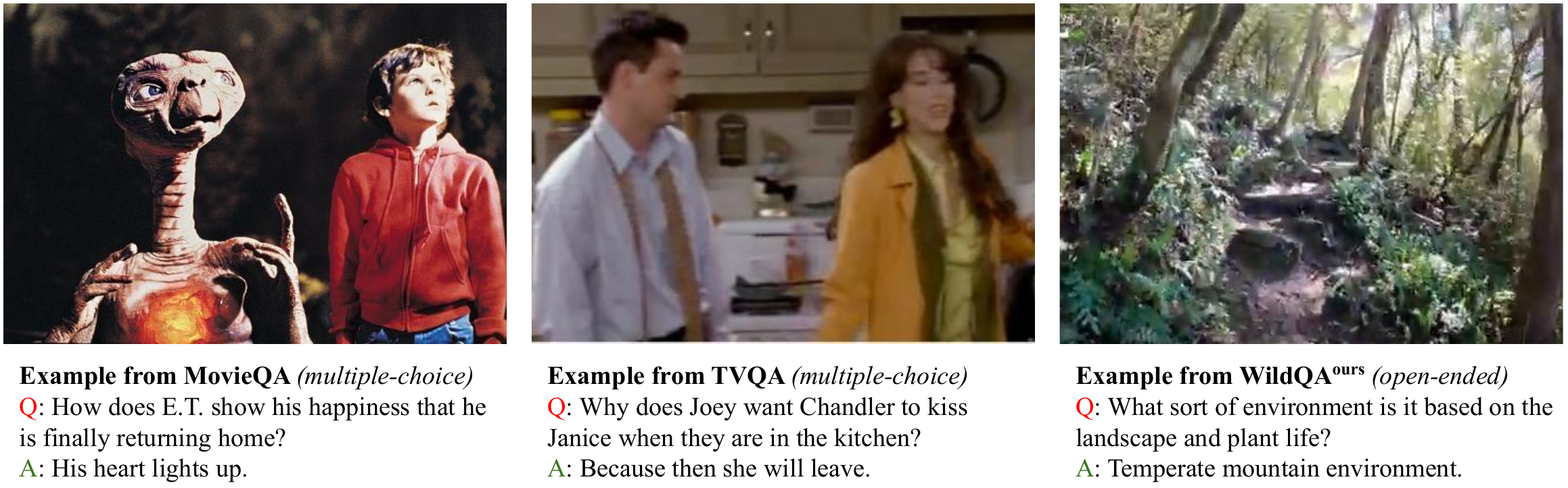}
\caption{Examples from MovieQA~\cite{movieqa}, TVQA~\cite{tvqa}, and our WildQA dataset. The previous datasets mostly focus on human interactions in a multiple-choice setting, while ours focus on scenes recorded in the outside world in an open-ended setting. We only list a single answer here for illustration purposes.}
\label{fig:comparison-example}
\end{figure*}

Video understanding plays an important role in the development of intelligent AI systems, as it enables the effective processing of different modalities of information \cite{li2020unimo}. Various tasks have been proposed to examine the ability of models' to understand videos, including video question answering (Video QA), video captioning, and fill-in-the-blank tasks~\cite{xu2017video,tran2016videomcc,castro2021fill}. Recent years have witnessed significant progress in video understanding, including new benchmarks~\cite{movieqa,grauman2021ego4d} as well as advanced sophisticated benchmarksmodels~\cite{jin2019multi,radford2021learning}.

There are however several drawbacks associated with existing video understanding research. First, existing video understanding benchmarks focus on common human activities as typically appearing in cooking videos ~\cite{zhu2017uncovering} or in movies~\cite{movieqa}, leading to a limited set of video domains. Second, most video understanding benchmarks adopt a multiple-choice format, where models select an answer from a set of candidates~\cite{jang2017tgif,castro2020lifeqa}. Models trained under such a setting cannot be used in real-life applications because candidate answers are not provided~\cite{castro2021fill}. Third, videos included in existing benchmarks are typically short~\cite{kim2016pororoqa}, and the performance of models on longer videos is not well studied.

We address these challenges in our dataset construction process. First, we propose the \wildqa{} dataset in which we collect ``in the wild'' videos that are recorded in the outside world, going beyond daily human activities. \Cref{fig:comparison-example} shows the difference between the \wildqa{} dataset and previous question answering datasets. Second, we adopt the challenging answer generation approach, aiming to build a system that can answer questions with an open-ended answer, rather than selecting from a predefined set of candidate answers. Third, the average video length in our dataset is one minute, longer than the video clips in most of the existing datasets in \cref{tab:dataset-comparison}, which presents a novel challenge for video understanding algorithms.


Using the \wildqa{} dataset, we address two main tasks. First, we address the task of video question answering (\textbf{Video QA}), aiming to generate open-ended answers. Second, we introduce the task of retrieving visual support for a given question and answer (\textbf{Video Evidence Selection}). 
Finding the relevant frames in a video for a given question-answer pair can help a system in its reasoning process, and is in line with ongoing efforts to build interpretable models~\cite{jacovi2020towards}. For each of these two tasks, we evaluate several baseline models, including multi-task models that combine the two tasks together. 
\Cref{fig:evidence-finding-example} shows an example from our dataset, including an example of a question, answer, and supporting video evidence.




To summarize, the main contributions of this paper are:

\begin{enumerate}[noitemsep,topsep=0pt]
\item We propose \wildqa{}, a multimodal video understanding dataset where video scenes are recorded in the outside world.
\item We propose two tasks for \wildqa{}: Video QA and Video Evidence Selection, aiming to build more interpretable systems.
\item We test several baseline models; experimental results show that our dataset poses new challenges to the vision and language research communities.
\end{enumerate}

\section{Related Work}

\paragraph{Multimodal Question Answering.}


Two popular and representative tasks are Visual Question Answering (Visual QA) on images, and Video Question Answering (Video QA) on videos. Visual QA has attracted attention for a long time~\cite{malinowski2015multiworld, zhang2016yin, ren2015exploring, zhu2016visual7w}. Recently, much progress has been made in Video QA. Researchers proposed various datasets such as TVQA that contain videos from movies or TV series~\cite{movieqa,tvqa,lei2019tvqa+} or videos from the Internet spanning from YouTube videos to Tumblr GIFs~\cite{zeng2017leveraging,youtube2text-qa,jang2017tgif,yu2019activitynet}. Other datasets such as MSVD-QA~\cite{xu2017video} contain videos from the existing corpus~\cite{chen2011collecting} or cartoon videos~\cite{kim2016pororoqa}. Recent Video QA datasets have stronger focuses such as temporal relations~\cite{mun2017marioqa}, multi-step and non-factoid answers~\cite{colas2019tutorialvqa}, natural interactions~\cite{zadeh2019social}, characters in the video~\cite{choi2020dramaqa}, question answering in real life~\cite{castro2020lifeqa}, incorporating external knowledge~\cite{garcia2020knowit}, and videos recorded from the egocentric view~\cite{fan2019egovqa,grauman2021ego4d}. To the best of our knowledge, we are the first to collect videos from the outside world. 

Researchers have also developed various methods to handle the Video QA task, including joint reasoning of the spatial and temporal structure of a video~\cite{zhao2017video, gao2019structured,huang2020location,jiang2020divide}, integrating memory to keep track of past and future frames~\cite{kim2017deepstory,gao2018motion,zhao2018open,fan2019heterogeneous,yu2020long}, various attention mechanisms~\cite{zhu2017uncovering,zhang2019frame,li2019beyond,yu2019compositional,kim2018multimodal,jin2019multi}, and others. Recently, pre-trained models have proved to be useful in various visual and language tasks~\cite{radford2021learning,chen2020uniter, zellers2021merlot}. However, the pre-trained visual and language models are typically encoder-only and cannot generate an answer in natural language on their own. Thus, such pre-trained encoder-only models do not fit into the open-end video question answering setting in our task.

Additionally, previous work has also investigated various reasoning tasks in a multimodal setting~\cite{gao2016physical, yang2018commonsense, gao2018what, zellers2019vcr}.
Although it is not our focus, some questions in our dataset require a certain level of reasoning ability. Moreover, since our dataset is created by domain experts, there is domain knowledge involved in the questions as well.


\paragraph{Moment Retrieval.}

Moment Retrieval
is the task of retrieving a short moment from a large video corpus given a natural language query~\cite{escorcia2019temporal,lei2020tvr}. Researchers have proposed or adapted various datasets for this task~\cite{krishna2017dense,anne2017localizing,gao2017tall,lei2020tvr}. The task of retrieving relevant parts in the video given the question (Video Evidence Selection) in our proposed dataset is akin to Moment Retrieval. However, moment retrieval focuses on retrieving the part of videos that the question describes, while Video Evidence Selection is to find parts of videos that can support the answer to the questions as shown in~\cref{fig:evidence-finding-example}. Prior work such as Tutorial-VQA~\cite{colas2019tutorialvqa} also adopt the setting of providing parts of the videos as answers to the question, but they did not include any text answers in their dataset. 

\paragraph{Few-shot Learning.}

Recently, there is a trend to evaluate neural models in a few-shot learning setting~\cite{huang2018natural,mukherjee2020uncertainty,sun2020neural,li2021few,lee2021kaggledbqa,pfeiffer2021xgqa}, where the model is tuned with a small portion of the data and tested against the rest. We adopt the few-shot learning setting for our dataset for both Video QA and Video Evidence Selection.


\section{WildQA Dataset}
\label{sec:wildqa-dataset}

\paragraph{Video Selection and Processing.} Following~\citet{zadeh2019social,castro2020lifeqa}, we start by collecting videos from YouTube. First, we identify five domains that primarily consist of outdoor scenes and are representative for the outside world, namely, \fdn{Agriculture}, \fdn{Geography}, \fdn{Human Survival}, \fdn{Natural Disasters}, and \fdn{Military}. We then manually collected videos from relevant YouTube channels for each domain.

Because the raw videos can be as long as an hour, we split the raw videos into short clips using PySceneDetect,\footnote{PySceneDetect uses the OpenCV~\cite{opencv_library} to find scene changes in video clips (\url{py.scenedetect.com}).} and concatenate these short clips so that the output video is approximately one minute. We use the output videos for the annotation process described below. More details for the video selection and video processing steps are discussed in \cref{appendix-subsec:video-selection}.


\paragraph{Question, Answer, and Evidence Annotation.} 

\begin{figure}
\centering
\includegraphics{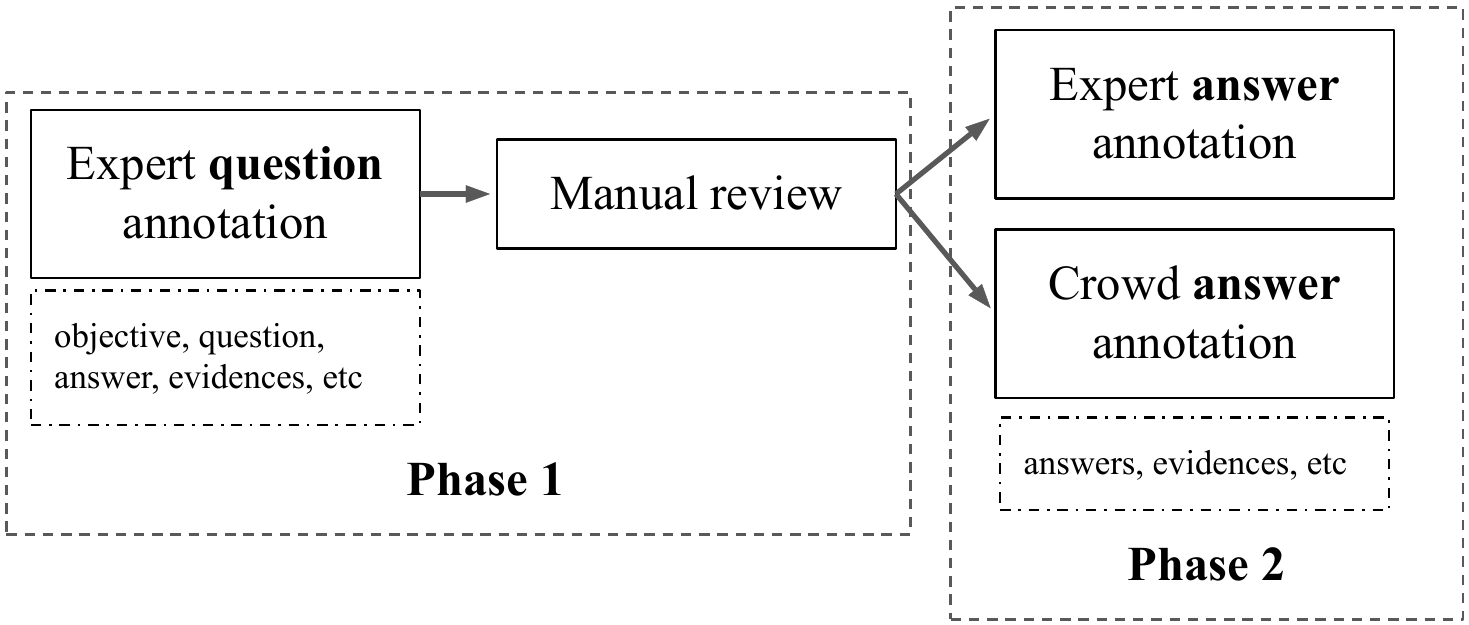}
\caption{The two phases of data annotation.}
\label{fig:annotation-phases}
\end{figure}

There are two phases in our annotation process, as shown in \cref{fig:annotation-phases}. In \textbf{Phase 1}, annotators watch the video clips and come up with a hypothetical motivation. They ask one or more \textbf{question}s and provide an \textbf{answer} to each of the questions they ask. Annotators are also instructed to provide all the relevant parts in videos as pieces of \textbf{evidence} to support the answer to their question. After this step in the data collection, three of the authors of this paper manually review all the question-answer pairs for quality purposes. Next, in \textbf{Phase 2}, we collect more \textbf{answer}s and \textbf{evidence}s for each question from Phase 1. Over the entire annotation process, annotators spent a total of \textbf{556.81 annotation hours}, split into 77.05 hours in Phase 1 and 479.76 in Phase 2. \Cref{appendix-subsec:annotation-instructions,appendix-subsec:annotation-interface,appendix-subsec:question-answer-corrections} present the annotation instructions, annotation interfaces, and reviewing process for question-answer pairs, respectively.

Because we want to collect questions that domain experts are interested, as opposed to arbitrary questions, domain experts carry out the Phase 1 annotations. To demonstrate the quality difference of questions collected from domain experts versus non-experts, we conduct a pilot study. \cref{appendix-subsec:pilot-study,appendix-subsec:annotator-information} discuss the pilot study and the annotators' expertise, respectively.

\paragraph{Dataset Statistics.}

\begin{table}
\small
\begin{tabular}{lrr}
\toprule
Domain           & \multicolumn{1}{l}{video count} & \multicolumn{1}{l}{question count} \\
\midrule
\dn{Agriculture}      & 85                              & 109                                \\
\dn{Human Survival}   & 95                              & 309                                \\
\dn{Natural Disaster} & 70                              & 187                                \\
\dn{Geography}        & 46                              & 110                                \\
\dn{Military}         & 73                              & 201                                \\
\hline 
Total            & 369                             & 916                               \\
\bottomrule
\end{tabular}
\caption{Video and question count for each domain.}
\label{tab:VQ-count}
\end{table}

\begin{table}
\small
\begin{tabular}{lc}
\toprule
Videos              & 369                                 \\
Duration (in seconds)  & 71.22 $\pm$ 26.47 \\
\midrule
Questions           & 916                                 \\
Question per video  & 2.48 $\pm$ 1.38                          \\
Question length (\#tokens)    & 7.09 $\pm$ 2.60                          \\
\midrule
Answer per question & 2.22 $\pm$ 0.69                  \\
Answer length (\#tokens)      & 9.08 $\pm$ 8.15                 \\
\midrule
Evidence per answer & 1.18 $\pm$ 0.80                  \\
Evidence length (s)    & 9.64 $\pm$ 10.96                \\
\bottomrule
\end{tabular}
\caption{Dataset statistics for WildQA.}
\label{tab:anno-info}
\end{table}

\begin{figure}
\centering
\includegraphics[width=0.9\textwidth]{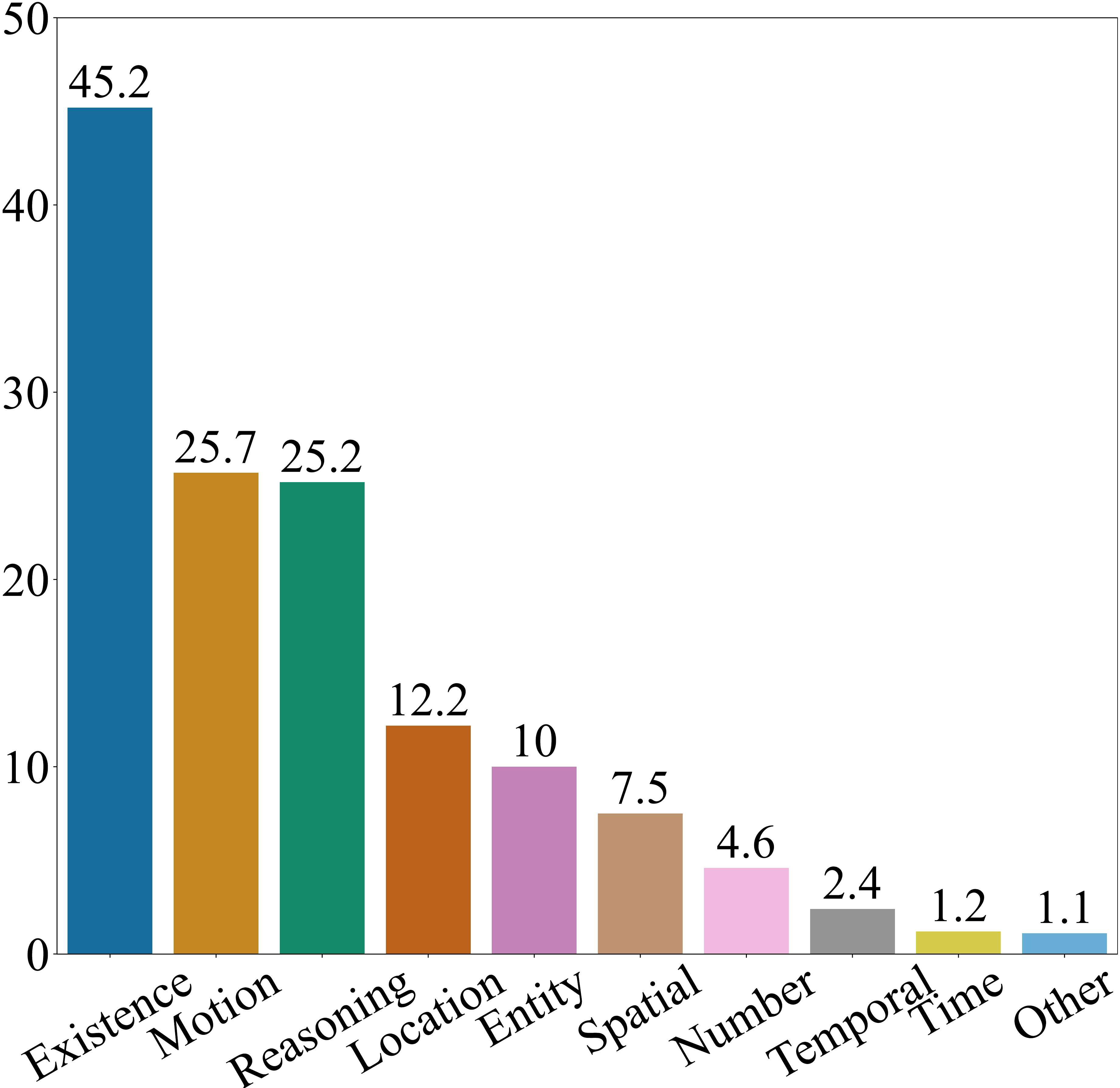}
\caption{Percentage distribution of question types. Because one question might be classified into multiple categories, the scale summation is larger than $100\%$.}
\label{fig:question-type}
\end{figure}

\Cref{tab:VQ-count,tab:anno-info} present statistics of the videos and associated questions for each of the five domains, along with other relevant statistics. \Cref{fig:question-type} shows the distribution of question types. \Cref{appendix-subsec:dataset-analysis} discusses more statistics.


\begin{figure}
\centering
\includegraphics{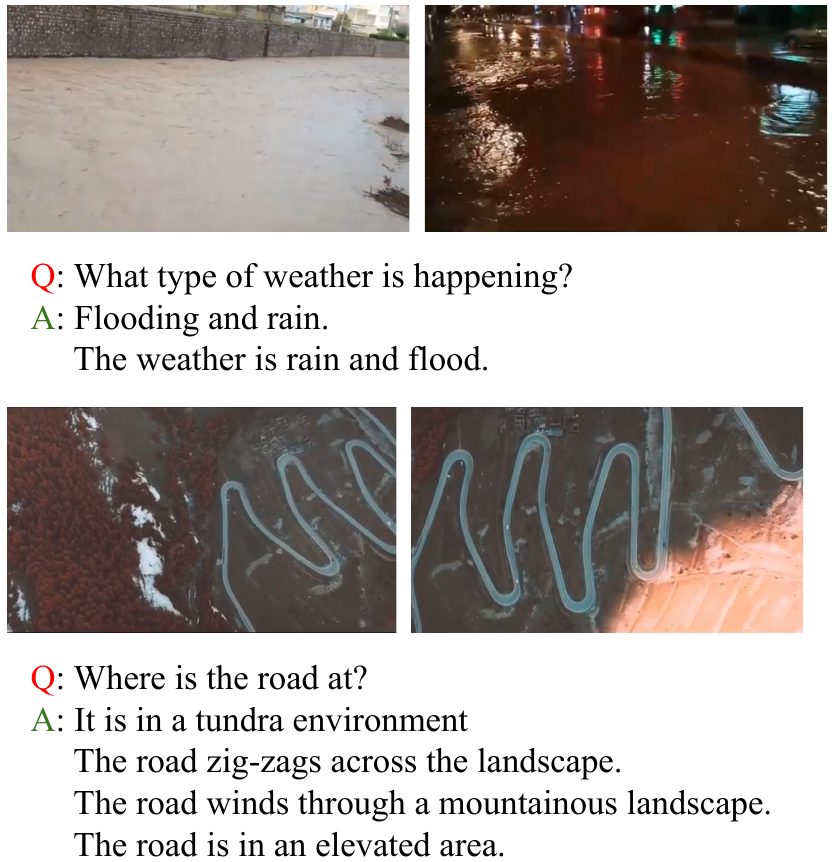}
\caption{Examples of questions (Q) and answers (A) from WildQA. The first answer is collected during Phase 1 of the annotation process; all remaining answers are collected in Phase 2.  More analyses in \cref{appendix-subsec:dataset-analysis}.}
\label{fig:more-dataset-examples}
\end{figure}

\paragraph{Dataset Comparison.}

\Cref{tab:dataset-comparison} shows the comparison between \wildqa{} and other existing datasets.

\begin{table*}
\footnotesize
\begin{tabular}{lccrrrcc}
\toprule
Dataset        & Domain                            & VE? & \begin{tabular}[c]{@{}l@{}}\#Videos\end{tabular} & \# Q   & \begin{tabular}[c]{@{}l@{}}Avg \\ dur. (s)\end{tabular} & Annotation & QA Task \\
\midrule
MovieQA \scriptsize{\cite{movieqa}} & Movies                      & $\surd$ & 6.7K                                                    & 6.4K   & 203                                                   & Manual        & MC      \\

VideoQA (FiB) \scriptsize{\cite{zhu2017uncovering}} & Cooking, movies, web       &     & 109K                                                    & 390K   & 33                                                & Automatic        & MC      \\
MSRVTT-QA \scriptsize{\cite{xu2017video}} & General life videos               &     & 10K                                                     & 243K   & 15                                                    & Automatic        & OE      \\

MovieFIB \scriptsize{\cite{moviefib}} & Movies                  &     & 128K                                                    & 348K   & 5                                                     & Automatic        & OE      \\

TVQA \scriptsize{\cite{tvqa}} & TV shows    & $\surd$ & 21.8K                                                   & 152K   & 76                                                    & Manual        & MC      \\
ActivityNet-QA \scriptsize{\cite{yu2019activitynet}} & Human activity                    &     & 5.8K                                                    & 58K    & 180                                                     & Manual        & OE      \\
TVQA+ \scriptsize{\cite{lei2019tvqa+}} & TV shows    & $\surd$   & 4.2K                                                    & 29.4K  & 60                                                      & Manual        & MC, ES  \\
KnowIT VQA \scriptsize{\cite{garcia2020knowit}} & TV shows      &     & 12K                                                     & 24K    & 20                                                      & Manual        & MC      \\
LifeQA \scriptsize{\cite{castro2020lifeqa}} & Daily life         &     & 275                                                     & 2.3K   & 74                                                      & Manual        & MC      \\

TutorialVQA \scriptsize{\cite{colas2019tutorialvqa}} & Instructions              & $\surd$   & 76                                                      & 6.2K   & --                                                      & Manual        & ES      \\

NExT-QA \scriptsize{\cite{nextqa}} & Daily life            &     & 5.4K                                                    & 52K    & 44                                                      & Manual        & MC, OE   \\

FIBER \scriptsize{\cite{castro2021fill}} & Human actions        &     & 28K                                                     & 2K     & 10                                                      & Manual        & OE      \\

\midrule
WildQA         & \textbf{In-the-wild}        & $\surd$   & 369                                                     & 916    & 71.2                                                   & Manual        & OE, \textbf{ES}  \\
\bottomrule
\end{tabular}
\caption{Comparison between our \wildqa{} and other existing datasets. \textbf{VE?}: Whether the dataset provides ``Video Evidences''?; \textbf{MC}: ``Multiple Choice'' question answering; \textbf{OE}: ``Open Ended'' question answering; \textbf{ES}: ``Evidence Selection''. We adapt the comparison table from~\citet{zhong2022video}.}
\label{tab:dataset-comparison}
\end{table*}

\section{Video Question Answering}
\label{sec:video-question-answering}

Following~\citet{xue2017unifying}, we adopt \textbf{free-form open-ended} video question answering for our video question answering (Video QA) task. Given a question \textbf{q} and a video \textbf{v}, the task is to generate an answer \textbf{a} in natural language.

We adopt a \textbf{few-shot learning} setting on our dataset, where models are fine-tuned on question-answer pairs corresponding to $30\%$ of the videos for each domain. The tuned models are tested on data for the remaining $70\%$ videos. The reason is that the time to annotate $30\%$ of the data is around 23 hours, during which there are around 50 data points annotated for each domain, which is acceptable. We hypothesize that it is realistic to have such a setting because the potential end-users could spend around a day or two collecting data, and we can then quickly tune a model using it. Moreover, no repeated videos appear in different splits, following~\citet{tvqa}. We end up having 264 question-answer pairs for 108 videos in our dev set and 652 pairs for 261 videos in the test set. We adopt BLEU~\cite{papineni2002bleu} and ROUGE~\cite{lin2004rouge} as the metrics to measure the quality of the generated answer. We run each model 3 times and report the scores of mean $\pm$ standard deviation in \cref{tab:few-shot-qa-results-10-epochs}.


\subsection{Baselines}

\paragraph{Human Baselines.} We report the average BLEU and ROUGE scores by leaving one annotator out in \cref{tab:few-shot-qa-results-10-epochs} (\fmn{Human}).

\paragraph{Text-only Models.}
We implement several baselines that only use the question-answer pairs in the dev set. \fmn{Random} randomly chooses answers from the dev set.
\fmn{Common} always predicts the most common answer in the dev set; \fmn{Closest} employs embedding produced by a pretrained \texttt{roberta-base} model~\cite{liu2019roberta}. In the inference, \fmn{Closest} retrieves the answers for the dev set question whose embedding has the highest cosine similarity to the test question. We also fine-tune T5~\cite{raffel2019exploring} using question-answer pairs from the dev set (\fmn{T5 T}).


\paragraph{Text + Visual Models.}
Following~\citet{castro2021fill}, we concatenate the text features with the visual features and input the concatenated features to the T5 model (\fmn{T5 T+V}). We extract I3D~\cite{carreira2018quo} video features and take one feature per second. 


\paragraph{Multi-task Learning.}

Multi-task learning has proved to be successful in various domains~\cite{collobert2008unified,deng2013new,girshick2015fast}. Following~\citet{Caruana93multitasklearning:}, we train \fmn{Multi$_\text{T+V,SE}$} which combines \mn{T5 T+V} and \mn{T5 SE} (the Video Evidence Selection model described in \cref{sec:video-evidence-selection}) with a shared T5 encoder between the tasks of Video Question Answering and Video Evidence Selection. We also train \fmn{Multi$_\text{T+V,IO}$} which combines \mn{T5 T+V} and \mn{T5 IO} (another Video Evidence Selection model described in \cref{sec:video-evidence-selection}) in a similar way. The loss function during the fine-tuning is:

\begin{equation}
\label{eq:multi-task-learning}
    L = \alpha L_1 + \beta L_2
\end{equation}
where $L_1, L_2$ are the losses for Video Question Answering and Video Evidence Selection, respectively; $\alpha, \beta$ are the weights for the two tasks. The selection process behind the values of $\alpha$ and $\beta$ are presented in \cref{sec:details-of-multi-task-learning}.

\begin{table}
\small
\begin{tabular}{lrrrrrr}
 \toprule
 Model name & ROUGE-1 & ROUGE-2 & ROUGE-L \\
 \midrule
 \mn{Random} & 5.0 $\pm$ 0.2&0.5 $\pm$ 0.1&4.9 $\pm$ 0.2\\
 
 \mn{Common}  & 10.6 $\pm$ 0.0 & 0.0 $\pm$ 0.0 & 10.6 $\pm$ 0.0  \\
 
 \mn{Closest}  & 19.5 $\pm$ 0.0 & 6.2 $\pm$ 0.0 & 18.7 $\pm$ 0.0 \\
 
\hdashline

 \mn{T5 T$^\text{0-shot}$}  & 0.8 $\pm$ 0.0 & 0.0 $\pm$ 0.0 & 0.8 $\pm$ 0.0 \\
 
 \mn{T5 T} & 33.8 $\pm$ 0.2& 17.7 $\pm$ 0.1& 32.4 $\pm$ 0.3 \\
 
 \mn{T5 T+V}   & 33.1 $\pm$ 0.3 & 17.3 $\pm$ 0.4 & 31.9 $\pm$ 0.2  \\

  \mn{Multi$_\text{T+V,IO}$}&\textbf{34.0} $\pm$ \textbf{0.5}&\textbf{18.8} $\pm$ \textbf{0.7}&\textbf{32.8} $\pm$ \textbf{0.6}\\
  
  \mn{Multi$_\text{T+V,SE}$} &33.8 $\pm$ 0.8&18.5 $\pm$ 0.7&32.5 $\pm$ 0.8\\
  
  \midrule 
  \mn{Human} & 40.8 $\pm$ 0.0 & 18.1 $\pm$ 0.0 & 36.3 $\pm$ 0.0 \\
 \bottomrule
\end{tabular}
\caption{ROUGE scores for the task of Video Question Answering. For comparison, we test the out-of-box T5 model under the zero-shot setting (\mn{T5 T$^\text{0-shot}$}).}
\label{tab:few-shot-qa-results-10-epochs}
\end{table}

\subsection{Results}
\label{subsec:qa-results}

\Cref{tab:few-shot-qa-results-10-epochs} reports F1 scores of ROUGE-1 (R1), ROUGE-2 (R2), and ROUGE-L (RL) for our baseline models. For comparison, we also test the out-of-box T5 model on our test split under the zero-shot setting (\fmn{T5 Text}$^\text{0-shot}$ in \cref{tab:few-shot-evidence-results-10-epochs}). 

T5-based models significantly outperform the random baselines as well as the out-of-box T5 model, which suggests that the T5-based models acquire certain levels of question-answering ability in the tuning stage. However, adding visual features does not improve the model's performance. This might be due to the \textit{challenges of attending to the visual features at the corresponding parts in the video}, because both models under multi-task learning outperform the text-only baseline, suggesting that attending to the correct part of the video helps the answer generation process.

All baseline models underperform human baselines on ROUGE scores, especially on ROUGE-1 and ROUGE-L scores, suggesting that there is room for improvement. However, the ROUGE-2 score for human annotators is low because although human annotators tend to use the same word to describe the object that appears in the video, there are large variations in terms of expressing the ideas of their answers. More discussions on the diversity of the answers are in \cref{appendix-subsec:dataset-analysis}.



\section{Video Evidence Selection}
\label{sec:video-evidence-selection}

Similar to \citet{colas2019tutorialvqa}, given a video \textbf{v} and a question \textbf{q}, the video evidence selection task consists of predicting $\{(\textbf{s}_1, \textbf{e}_1), (\textbf{s}_2, \textbf{e}_2), \ldots\}$, where $(\textbf{s}_i, \textbf{e}_i)$ represents the time for start \textbf{s} and end \textbf{e} of a singles span within the video \textbf{v}. We also adopt the few-shot learning setting as described in \cref{sec:video-question-answering} for the task of Video Evidence Selection. Similar to~\citet{deyoung2019eraser}, we design an Intersection-Over-Union
(IOU) metric borrowed from~\citet{everingham2010pascal}. We define IOU as follows: given two time spans in the video, IOU is defined as the length of their intersection divided by the length of their union. The prediction is counted as a match if it overlaps with any of the ground truth spans by more than the threshold (0.5, following~\citealp{deyoung2019eraser}). We use these partial matches to calculate an F1 score (IOU-F1 scores). As described in \cref{sec:video-question-answering}, we run each model three times and report the scores of mean $\pm$ standard deviation in \cref{tab:few-shot-evidence-results-10-epochs}.

\subsection{Baselines}

As described in \cref{sec:video-question-answering}, we compute the average IOU-F1 score on the annotations from one annotator against the remaining annotators; we denote this metric as \fmn{Human}. The \fmn{Random} baseline consists of randomly choosing the start and end of a part within the original video as evidence. Similar to the structure~\citet{devlin2019bert} experiment on SQuAD~\cite{rajpurkar2016squad}, we build \fmn{T5 SE}; here, we feed the concatenated question embeddings and I3D visual features to the T5 encoder, and the T5 encoder outputs a sequence of the encoded states. We treat the subsequence corresponding to the visual features as the encoded hidden sequence $T_m\in R^H$ for the video frames ($H$ denotes the dimension of the hidden sequence). We then multiply the sequence with two vectors $S, E \in R^H$. The $T_i$ and $T_j$ that maximize the likelihood are predicted as the \textbf{start and the end of the evidence}, respectively. During the training, we maximize their joint probability:

\[
P_i P_j = \frac{e^{S\cdot T_i}}{\sum_m e^{S\cdot T_m}} \frac{e^{E\cdot T_j}}{\sum_m e^{E\cdot T_m}}
\]

where $P_i$ and $P_j$ are the probability for the $i$ being the start and $j$ the end of the evidence, respectively.

Inspired by the Inside-Outside-Beginning (``IOB'') tagging scheme~\cite{ramshaw1999text}, we also formulate the evidence finding as a task of tagging whether a video frame is inside (``I'') the evidence, or outside (``O'') the evidence. We then build \fmn{T5 IO} by feeding the concatenated features to a T5 encoder. Similar to \mn{T5 Start End}, we have an encoded sequence of $T_m\in R^H$ corresponding to the video frames. We then multiply the sequence with a vector $L\in R^H$ and apply a sigmoid function to the multiplication result. The model predicts the frame as ``I'' if the value at the corresponding position is greater than or equal to $0.5$, otherwise it predicts ``O''. We test \mn{Multi$_\text{T+V,IO}$} and \mn{Multi$_\text{T+V,SE}$} described in \cref{sec:video-question-answering} on Video Evidence Selection as well.

\subsection{Results}
\label{subsec:evidence-results}

\begin{table}
\small
\begin{tabular}{lr}
\toprule
Model name & IOU-F1 \\
\midrule
\mn{Random} &2.5 $\pm$ 0.3 \\
\hdashline
\mn{T5 IO}  &1.1 $\pm$ 0.2 \\
\mn{T5 SE} & \textbf{4.5} $\pm$ \textbf{0.8} \\
\mn{Multi$_\text{T+V,IO}$} &1.4 $\pm$ 0.3 \\
\mn{Multi$_\text{T+V,SE}$} &3.7 $\pm$ 2.4 \\
\hline
\mn{Human} & 18.37 $\pm$ 0.0 \\
\bottomrule
\end{tabular}
\caption{IOU-F1 scores for Video Evidence Selection.}
\label{tab:few-shot-evidence-results-10-epochs}
\end{table}

\Cref{tab:few-shot-evidence-results-10-epochs} shows the performance of the baseline models on the Video Evidence Selection task. All the baseline models perform significantly worse than the human annotators, and sometimes worse than the random baseline. This is understandable because selecting evidence from a long video can be difficult. Additionally, multi-task learning makes the model's performance worse. However, this could be due to the fact that the Video Evidence Selection itself is a hard task, and all the baseline models struggle with such a task. Although multi-task learning does not help Video Evidence Selection, as mentioned in \cref{sec:video-question-answering}, training with Video Evidence Selection does help Video QA. Thus, Video Evidence Selection is still an important task to improve a model's ability to answer questions. We include more ablation studies in \cref{subsec: ablation-study-on-video-evidence-selection}.





\begin{table}
\small
\begin{tabular}{lrr}
 \toprule
 Type & R1 & IOU-F1 \\
 \midrule
\textit{Existence} &\textbf{33.3} $\pm$ \textbf{0.3} & \textbf{5.3} $\pm$ \textbf{0.3}\\
\textit{Motion} & 32.8 $\pm$ 0.6 & 3.1 $\pm$ 2.0\\
\textit{Reasoning} & 33.3 $\pm$ 0.4&3.1 $\pm$ 1.3\\
\textit{Location} & \textbf{26.2} $\pm$ \textbf{10.7}&\textbf{4.4} $\pm$ \textbf{1.4}\\
\textit{Entity} & 33.2 $\pm$ 0.7&5.2 $\pm$ 0.7\\
\textit{Spatial} & \textbf{32.2} $\pm$ \textbf{0.6}&\textbf{2.4} $\pm$ \textbf{1.7}\\
\textit{Number} & 33.8 $\pm$ 0.4&4.5 $\pm$ 0.7\\
\textit{Temporal} & 33.8 $\pm$ 0.6&3.8 $\pm$ 0.5\\
\textit{Time} & 33.1 $\pm$ 0.8&5.7 $\pm$ 1.0\\
\textit{Other} & 33.2 $\pm$ 0.6&5.3 $\pm$ 0.9\\
 \bottomrule
\end{tabular}
\caption{\mn{Multi$_\text{T+V,SE}$} performance on different question types for Video QA (ROUGE-1) and for Video Evidence Selection (IOU-F1).}
\label{tab:model-performance-type-performance-rouge-1-iou}
\end{table}

\section{Analysis and Discussion}
\label{sec:analysis-and-discussion}

\paragraph{Model Performance v.s. Question Types.}

\Cref{tab:model-performance-type-performance-rouge-1-iou} shows \mn{Multi$_\text{T+V,SE}$}'s performance on different question types for Video QA and Video Evidence Selection respectively. Other ROUGE scores for Video QA follow similar trends as shown in \cref{fig:model-performance-type-performance}. According to \cref{tab:model-performance-type-performance-rouge-1-iou}, the model achieves good ROUGE-1 scores for Video QA when the model has a good IOU-F1 score for Video Evidence Selection such as its performance on \textit{Existence}. The model has the highest ROUGE-1 variation on \textit{Location} question types, with a relatively large variation for IOU-F1. The model's ROUGE-1 score on \textit{Spatial} questions is relatively low, with the lowest IOU-F1 score. \mn{Multi$_\text{T+V,SE}$} excels at question type \textit{Entity} and \textit{Existence} with relatively high IOU-F1 scores. One possible explanation could be that the average length of the answers generated for \textit{Entity} and \textit{Existence} are around eight tokens, which might be easier for the model to ground to the relevant part in the video.

Interestingly, even if the answers have similar lengths, the model struggles on \textit{Motion} questions (with a relatively low IOU-F1 score). A possible reason could be that this type of questions provide a very abstract description of the action, which makes the model hard to attend to the relevant part of the video. For instance, an example of a \textit{Motion} question is \textit{``Are there any structure or natural features being affected?''}. To attend to the corresponding period in the video, the model needs to understand the word ``affected'' and the objects that are actually affected, which can be very difficult. The model also struggles to attend to the correct places in the video for the \textit{Spatial} type of question. This might be because there is more than one entity in \textit{Spatial} type of questions, and the model needs to locate all the objects appearing in various parts of the video, which is similarly complex. For instance, for the question \textit{``What effects did the weather have?''}, the model needs to attend to \textit{``debris in the air''}, \textit{``truck turnover''} and \textit{``destruction of buildings''}. For \textit{Location} type of questions such as \textit{``What sorts of terrain is the vegetation present in?''}, it might be difficult to attend to all the terrains of \textit{``forest''}, \textit{``plateaus''}, \textit{``mountainous''}, \textit{``valleys''}, and \textit{``arboreal''} and to include them in the answer.


\begin{table}
\small
\begin{tabular}{lrrr}
 \toprule
 Model name & R\@1 & R\@2 & R\@L \\
 \midrule
 
 \mn{T5 T}$^\text{0-shot}$  & 0.8 $\pm$ 0.0 & 0.0 $\pm$ 0.0 & 0.8 $\pm$ 0.0 \\
 
 \mn{T5 T}$_\text{TVQA}^\text{0-shot}$ &9.1 $\pm$ 0.0&1.2 $\pm$ 0.0&8.8 $\pm$ 0.0\\

\mn{T5 T}$_\text{TVQA,ours}$ &32.4 $\pm$ 0.2&17.5 $\pm$ 0.2&31.6 $\pm$ 0.2\\
  
 \mn{T5 T}$_\text{ours}$  & \textbf{33.8 $\pm$ 0.2} & 17.7 $\pm$ 0.1 & 32.4 $\pm$ 0.3 \\

 \hdashline
 
 \mn{T5 T+V}$_\text{TVQA}^\text{0-shot}$ &20.3 $\pm$ 0.0&8.1 $\pm$ 0.0&20.1 $\pm$ 0.0\\
 \mn{T5 T+V}$_\text{ours}$  & 33.1 $\pm$ 0.3 & 17.3 $\pm$ 0.4 & 31.9 $\pm$ 0.2  \\
 
  \mn{T5 T+V}$_\text{TVQA,ours}$ &33.7 $\pm$ 0.2&\textbf{18.3 $\pm$ 0.1}& \textbf{32.6 $\pm$ 0.1}\\
 
 
 
 \bottomrule
\end{tabular}
\caption{ROUGE scores for the task of Video Question Answering for few-shot learning setting (the standard setting in our WildQA dataset introduced in \cref{sec:video-question-answering}) and zero-shot learning setting (``0-shot'' in the superscript). Subscript ``TVQA'' means pre-training on the TVQA~\cite{tvqa} dataset; subscript ``TVQA,ours'' means first pre-training the model on TVQA, then tuning the model on our WildQA dataset; subscript ``ours'' means tuning the model directly on our WildQA dataset.}
\label{tab:pre-train-on-tvqa}
\end{table}

\begin{figure}
\centering
\includegraphics{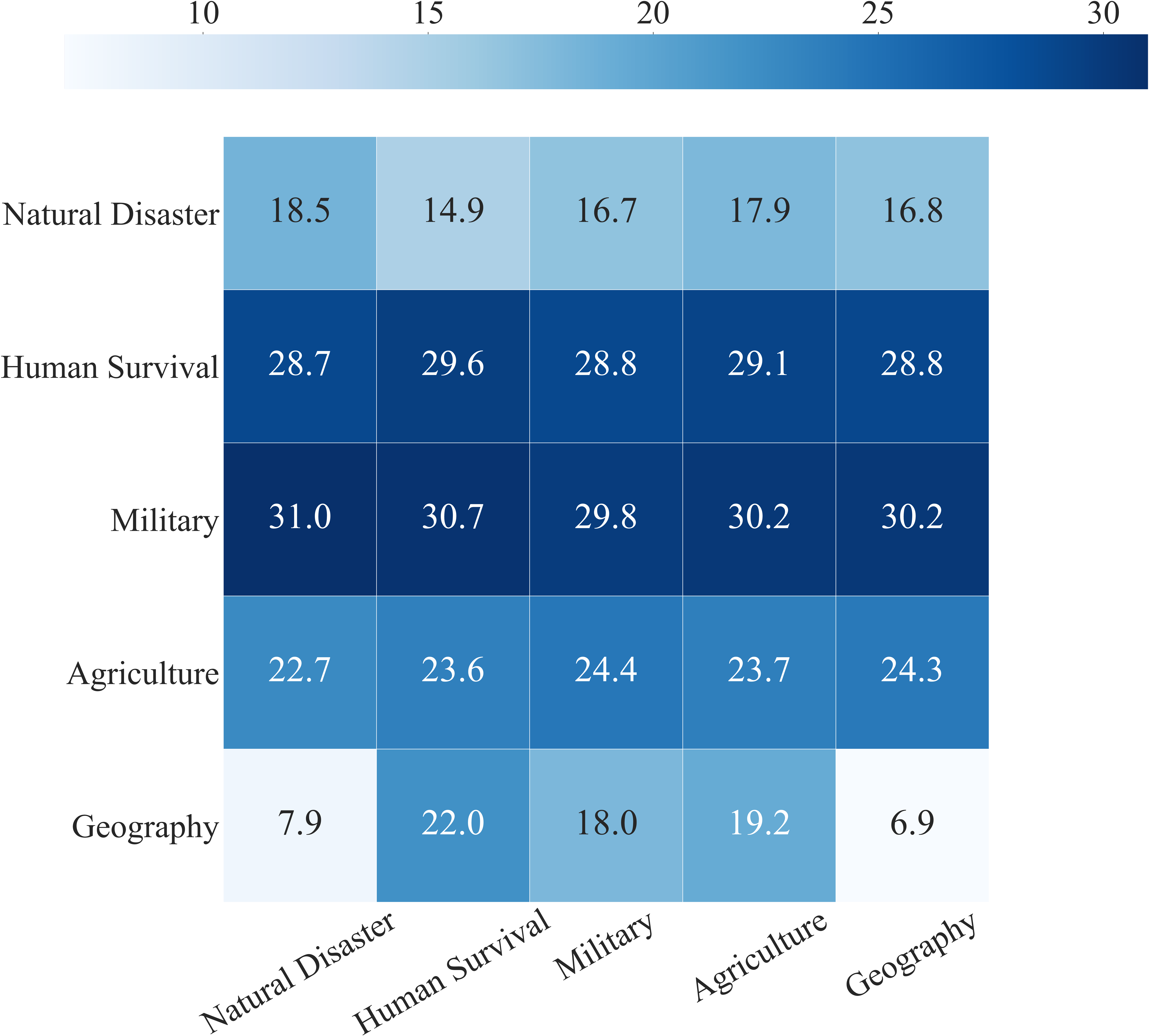}
\caption{\mn{Multi$_\text{T+V,SE}$} performance (ROUGE-1) for Video QA when tuned on a single domain (y-axis) and tested against each domain (x-axis). The performances by the rest metrics for Video QA resemble the pattern here and are reported in \cref{appendix:experiment-results}.}
\label{fig:rouge-1-heatmap}
\end{figure}

\begin{figure}
\centering
\includegraphics{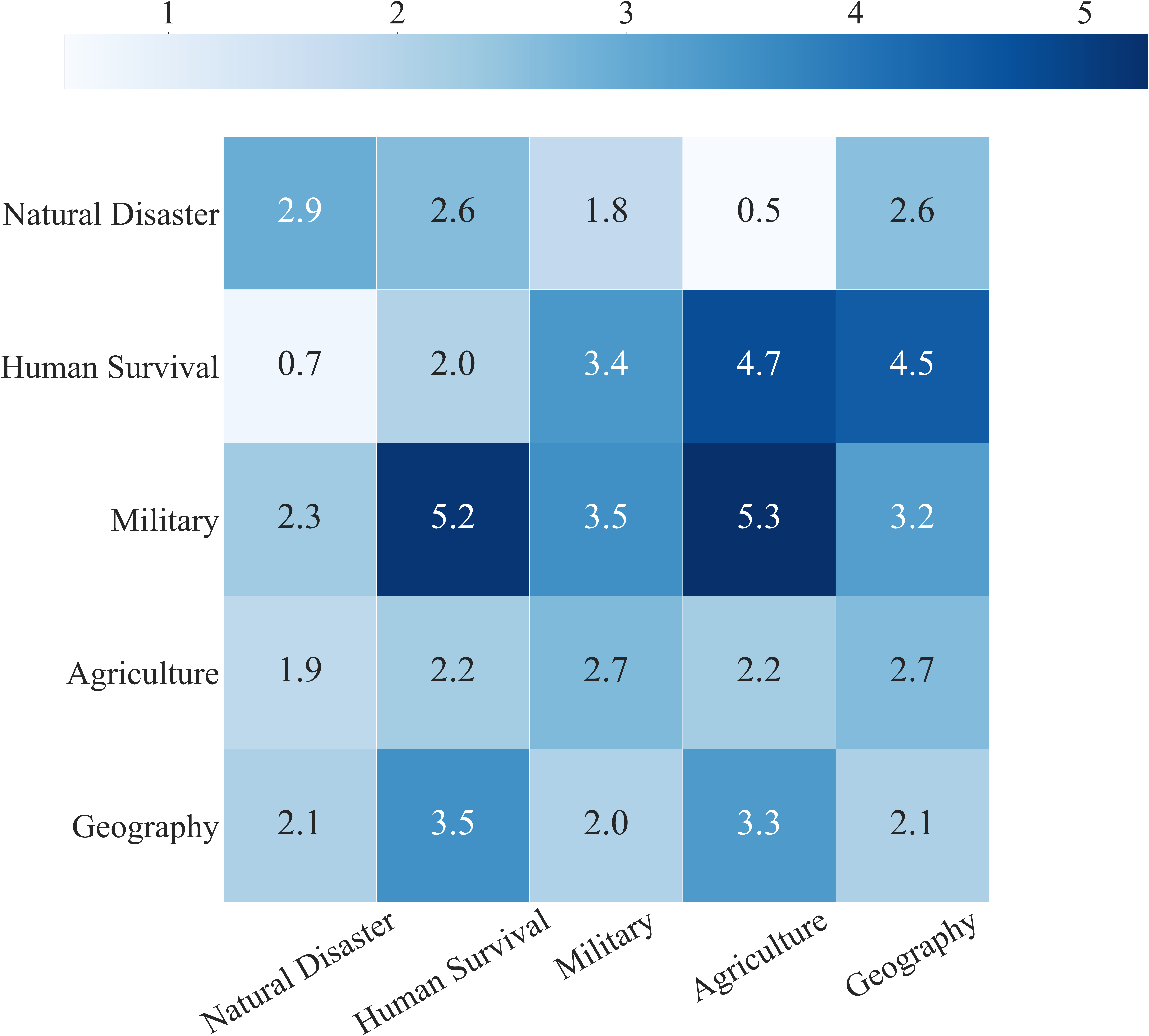}
\caption{\mn{Multi$_\text{T+V,SE}$} performance (IOU-F1) for Video Evidence Selection when tuned on a single domain (y-axis) and tested against each domain (x-axis).}
\label{fig:iou-f1-heatmap}
\end{figure}

\paragraph{Domain Adaptation.} Furthermore, we tune the \mn{Multi$_\text{T+V,SE}$} model on the dev set data from a single domain, and test it against data from other domains. \Cref{fig:rouge-1-heatmap,fig:iou-f1-heatmap} show the model's performance in different tuning and testing domains. Interestingly, the diagonal cells do not always have the darkest color, which indicates that inter-relations exist across domains. For instance, the model tuned on \dn{Geography} performs relatively better for Video QA on \dn{Human Survival} and \dn{Agriculture} rather than itself. This suggests that the questions and videos from \dn{Geography}, \dn{Agriculture}, and \dn{Human Survival} exhibit some similarity so that the model tuned on one domain can answer questions from the other domains relatively well. But answering questions from \dn{Geography} can introduce the domain knowledge, an example of the answer is \texttt{``Mountainous, temperate forest.''}, where \texttt{``temperate forest''} is one of the terminologies specific to \dn{Geography} domain. Training on these terminologies might confuse the model and hurt the performance. Thus, future research might be needed to study how to better incorporate domain knowledge into multimodal question answering.

As for Video Evidence Selection, the patterns generally resemble the pattern in \cref{fig:rouge-1-heatmap}, which means that in general, the model answers a question better if it can attend to the relevant part in the video. However, when tuned on \dn{Human Survival} and tested on \dn{Natural Disaster} the model performs relatively well on Video QA (with a 28.7 ROUGE-1 score) but less well on Video Evidence Selection (with a 0.7 IOU-F1 score). This might indicate that the model picks up some common patterns in the text rather than reasoning about the video and the question in an expected manner. 

\paragraph{Pre-training on Other Datasets.} We also pre-train the \mn{T5 T} and 
\mn{T5 T+V} using TVQA~\cite{tvqa}, a large-scale multimodal question answering dataset with videos from TV series. We report the zero-shot learning performances as well as the few-shot learning performances for \mn{T5 T} and \mn{T5 T+V} in \cref{tab:pre-train-on-tvqa}. We can see that pre-training on TVQA for text-only 
\mn{T5 T} does not help, which shows that the question styles in our dataset might be different from TVQA. For \mn{T5 T+V} which uses both text and visual features, pre-training on TVQA does help the model, which suggests that the pre-training helps the model take advantage of the visual features. \mn{T5 T+V} pre-trained on TVQA underperforms \mn{T5 T+V} trained together with \mn{T5 IO} (the \mn{Multi$_\text{T+V,SE}$} model) according to \cref{tab:few-shot-qa-results-10-epochs} and \cref{tab:pre-train-on-tvqa}, suggesting that attending to the relevant part in the video helps the model better than training the model on more data. However, pre-training the model on the TVQA dataset reduces the variance of model performance, which suggests that training the model with more data helps the model perform consistently.

\section{Conclusion}

In this paper, we introduced a new and challenging benchmark, \wildqa{}, to promote domain diversity for video understanding. Specifically, we focused on five domains that involve long videos recorded in the outside world, which can be  useful for applications  in these domains. Instead of the traditional multiple-choice setting for Video Question Answering, we proposed to generate open-ended answers. We believe open-end answer generation can help construct systems that can answer end users' questions in a more natural way. To help the model attend to the relevant parts in the videos, we also proposed the task of Video Evidence Selection. Through experiments, we showed the feasibility of these tasks, and also showed that jointly training for both Video Question Answering and Video Evidence Selection can improve the models' performance. In addition, we found it is easier to understand models' behavior by knowing which part of the video the model attends to when  answering a question. We believe that this is an important step towards a trustworthy, explainable multimodal system. The dataset is available at \url{https://lit.eecs.umich.edu/wildqa/}.

\section*{Acknowledgement}
We thank the anonymous reviewers for their constructive feedbacks. We thank Artem Abzaliev, Do June Min, and Oana Ignat for proofreading and suggestions. We thank William McNamee for the help with the video collection process, and all the annotators for their hard work on data annotation. We thank Yiqun Yao for the helpful discussions during the early stage of the project. 
This material is based in part upon work supported by the Automotive Research Center (``ARC''). Any opinions, findings, conclusions, or recommendations expressed in this material are those of the authors and do not necessarily reflect the views of ARC or any other related entity.

\section*{Ethical Consideration}

The videos we use are all publicly available on YouTube. The dataset includes a variety of domains, including videos in the \dn{Military} domain,  but we are ensuring that all the videos are only used for asking questions specific to the video content. Moreover, in our manual examination, we make sure that the question-answer pairs collected from our annotators are appropriate without any use of offensive language. 
\bibliography{main}
\bibliographystyle{acl_natbib}

\newpage
\appendix


\section{Annotation Details}
\label{appendix-sec:annotation-details}

\subsection{Video Selection and Processing}
\label{appendix-subsec:video-selection}

\paragraph{Video Selection.} For the video selection part, as mentioned in \cref{sec:wildqa-dataset}, first, we identify 5 domains,  \fdn{Agriculture}, \fdn{Geography}, \fdn{Human Survival}, \fdn{Natural Disasters}, and \fdn{Military}, to collect videos recorded in the outside world. We then identify eight (8) YouTube channels and crawl videos from those channels. During crawling, we manually substitute irrelevant videos such as advertisements with videos that contain scenes mostly recorded in the outside world from the same channel.

\paragraph{Video Processing.} As mentioned in \cref{sec:wildqa-dataset}, we clip the raw videos into short clips by PySceneDetect because the raw videos can be as long as an hour. We then concatenate these short clips so that the output video will be around 1 minute. \textbf{The output videos are used for the following annotation process.} We want to include longer videos because the videos recorded in the outside world usually contain less information compared to the videos about human interactions. Besides, if the concatenated video is at the end of the original video, it is allowed to be shorter than 1 minute. We select the concatenated videos that only contain scenes recorded in the outside world. If none of the concatenated videos satisfies, we manually clip the original videos to get an output video. 

\subsection{Annotation Instructions}
\label{appendix-subsec:annotation-instructions}

As mentioned in \cref{sec:wildqa-dataset}, we have 2 phases in our annotation process as shown in \cref{fig:annotation-phases}. In Phase 1, annotators come up with a hypothetical motivation, ask questions, and provide the corresponding answers with relevant parts of the video as evidence. Phase 2 is to collect answers and evidence for questions we collect in Phase 1. The following are the instructions for these two phases.

\paragraph{Instructions for Phase 1}

\begin{enumerate}[leftmargin=*]
    \item[] We need help for this Video QA task based on video content (including the audio).\\
    In this task, we suppose you can hypothetically send a robot to a place that you want, for many hours, so as to collect information that you need. In this hypothetical scenario, you have an objective that you want the robot to learn about. This robot can chart territory and is able to answer questions based on recorded videos. Therefore, after it comes back, you can ask questions to help you satisfy your objective, then this robot will provide you with answers, as well as video evidence clips to support the answers.\\
    In this task, to simplify, the provided videos represent places where you could potentially have sent the robot and are much shorter (a few minutes). Given a recorded video, please help us provide one hypothetical objective that makes sense with it, along with questions, answers, and evidence. Specifically, you should pretend to be both the information-seeker and the robot, which means that as the robot, you could watch the recorded video, and you should provide answers and video evidence clips; as the information-seeker, you have an objective, \textbf{not} watch the whole video (because of practical reasons), and you can only ask questions and receive answers and video evidence clips as feedback.
    \item Basic Instructions
    \begin{itemize}
        \item You will need to propose a hypothetical objective (or topic, intention, motivation), to motivate the questions, that makes sense for the given video.
        \item You will need to provide as many questions as you need (to satisfy your objective) with regard to the content in the videos and that seek to understand more about the proposed objective.
        \item You will first watch the video, but when you are providing the objective and questions, please pretend you \textbf{haven't} seen it before.
        \item You will need to provide at least one question for each video. \textbf{The more the better}.
        \item You will need to identify the source of your question (whether it is based on the visual scene or the audio) and classify your question accordingly.
        \item You will need to provide the correct answer to the question you asked, as supported by the content in the video.
        \item You will need to provide video evidence (video clip) to support your question and answer.
        \item If one video doesn't make sense at all, or there's no possible objective for this video that makes sense, please comment at the bottom of this annotation page (and fill in the mandatory fields for the corresponding video with placeholder values).
    \end{itemize}
    \item How To Propose Hypothetical Objective
    \begin{itemize}
        \item For each video, you need to come up with a hypothetical objective (or intention, motivation, topic) that makes sense for this video, and briefly explain it.
        \item Your questions should all relate to this objective.
        \item Example 1:
        \begin{itemize}
            \item Objective: I want to learn about the water in the territory.
            \item Question 1: How big is the lake?
            \item Question 2: Are there any boats in the lake?
            \item Question 3: Where is the river?
            \item ...
        \end{itemize}
        \item Example 2:
        \begin{itemize}
            \item Objective: people/life movement
            \item Question 1: Is there any sign that wildlife has passed this area?
            \item Question 2: How much traffic is there on the road?
            \item ...
        \end{itemize}
    \end{itemize}
    \item How To Ask Your Question
    \begin{itemize}
        \item Your question should relate to your proposed objective.
        \item For each video, after you finish one question, you could click the Add one more question for this video button to continue to provide another question for this video. On the contrary, if you want to delete one question, you could click the Delete this Question button.
        \item Ask one question at a time.
        \begin{itemize}
            \item E.g., \textit{"Are there any people? What are they doing?"} is not appropriate.
        \end{itemize}
        \item When you provide multiple questions for the same video, make sure these questions are \textbf{independently} asked.
        \begin{itemize}
            \item E.g., "What is growing on pine trees?" and "What is their color?" are not independent.
        \end{itemize}
        \item The answer should be derived from the video (visual or audio).
        \begin{itemize}
            \item E.g., \textit{"Why do they run every morning?"} is not a good question.
        \end{itemize}
        \item Ask from the 3rd person point of view.
        \begin{itemize}
            \item E.g., "What do \textit{we} have on this farm?" -> "What do \textit{They} have on this farm?"
        \end{itemize}
        \item Try to balance the questions such that the answers are not too repetitive (E.g., too many 'yes' answers).
        \item Ask questions matter-of-factly (as \textbf{objectively} as possible). Stick to what you can see or hear from the video.
        \begin{itemize}
            \item E.g., \textit{"Does it make people feel good here?"} is somehow subjective.
        \end{itemize}
        \item Don't ask questions about how's the video being recorded, the camera-person or the camera itself. Ask about the content itself. Ignore what the camera-person is doing.
        \begin{itemize}
            \item E.g., "What's the \textit{cameraman} doing?" / "How fast is the \textit{camera} moving?" are not good questions.
        \end{itemize}
    \end{itemize}
    \item How to identity the Question Category
    \begin{itemize}
        \item[] We have some basic categories: \textbf{Motion, Spatial Relationship, Temporal Relationship, Reasoning, Number, Entity, Existence, Time, Location, Other.}\\If your questions fall into \textbf{multiple categories}, please check all categories that apply.\\ Here are some example questions under each category:
        \item \textbf{Motion}: What is the group of soldiers doing?
        \item \textbf{Spatial Relationship}: What is driving beside the motorcycle?
        \item \textbf{Temporal Relationship}: What happens before the black smoke rises?
        \item \textbf{Reasoning}: What makes changing between targets possible for the missile?
        \item \textbf{Number}: How many fighters are flying?
        \item \textbf{Entity}: What is the target of the bullet?
        \item \textbf{Existence}: Is there a lake by the mountain?
        \item \textbf{Time}: How long can the missile fly?
        \item \textbf{Location}: Where is the tank?
        \item \textbf{Others}
    \end{itemize}
    \item How To Provide Correct Answer
    \begin{itemize}
        \item Your answer should be written as \textbf{full sentences} (at least one).
        \begin{itemize}
            \item E.g., \textit{"Left"} -> \textit{"The landspout bends toward the left."}
        \end{itemize}
        \item The answer should be derived from the video (visual or audio).
        \begin{itemize}
            \item E.g., "\textit{These plants are green because they contain chlorophyll.}" is not a good answer.
        \end{itemize}
        \item Provide answers matter-of-factly (as objectively as possible). Stick to what you can see or hear from the video.
        \begin{itemize}
            \item E.g., "\textit{beautiful}" is likely not a good word to use within an answer.
            \item E.g., "\textit{This takes some bravery to do.}" is somehow subjective.
        \end{itemize}
        \item Don't answer about how's the video being recorded, the camera-person, or the camera itself. Answer about the content itself. Ignore what the camera-person is doing.
        \begin{itemize}
            \item E.g., "There are two people, i.e. a running child, and the \textit{cameraman}." is not a good answer.
        \end{itemize}
        \item When you enter numbers, please enter digits instead of text.
        \begin{itemize}
            \item "\textit{Seventeen}" -> "\textit{17}"
        \end{itemize}
    \end{itemize}
    \item How to provide video evidence
    \begin{itemize}
        \item The video evidence consists of \textbf{all} the parts of the video that support the answer to your given question.
        \item You need to provide at least one video evidence clip (intervals within the video) for each question.
        \item You need to provide both the \textbf{start point and end point} for all the video evidence you identify in the video;
        \item You could use your mouse or ←/→ key to \textbf{click or drag the process bars} of start point and end point. When you click or drag the bar, the above video will change accordingly, so you could locate the points according to the video screen.
        \item For each video evidence clip, the end point should be \textbf{greater than zero}, and the end point should be greater or equal to the start point.
        \item The video evidence clips (the time gap between the start point and the end point) should be as short as possible.
    \end{itemize}
\end{enumerate}

\paragraph{Instructions for Phase 2}

\begin{enumerate}[leftmargin=*]
    \item[] We need help for this Video Question Answering task based on video content (including the audio).
    \item Basic Instructions
    \begin{itemize}[leftmargin=*]
        \item You will first watch the video, then answer the questions, each question in turn.
        \item You will need to provide at least one answer for each question (ignoring differences such as upper/lower case, or the article). \textbf{The more answers the better}, but every answer should be correct.
        \item You will need to identify the source of your answer (whether it is based on the visual scene or the audio).
        \item For each answer, you will need to provide video evidence (video clip) to support your answers. See below for additional information.
        \item If one video or question is not available, please comment at the bottom of this annotation page (and fill the mandatory fields for this video/question with placeholder values).
        \item There are five questions, you need to finish all five questions according to the content in the video (including audio).
    \end{itemize}
    \item How To Answer
    \begin{itemize}[leftmargin=*]
        \item Provide one or more answers for each question.
        \item Each answer should be written as full sentences (at least one).
        \begin{itemize}[leftmargin=*]
            \item E.g., "Left" -> "The landspout bends toward the left."
        \end{itemize}
        \item The answer should be derived from the video (visual or audio).
        \begin{itemize}[leftmargin=*]
            \item E.g., "These plants are green because they contain chlorophyll." is not a good answer.
        \end{itemize}
        \item Respond matter-of-factly (as objectively as possible). Stick to what you can see or hear from the video.
        \begin{itemize}[leftmargin=*]
            \item E.g., "beautiful" is likely not a good word to use within an answer.
            \item E.g., "This takes some bravery to do." is somehow subjective.
        \end{itemize}
        \item Answer in 3rd person point of view.
        \begin{itemize}[leftmargin=*]
            \item E.g.,"We raise cattle on this farm." -> "They raise cattle on this farm."
        \end{itemize}
        \item Don't answer about how's the video being recorded, the camera-person, or the camera itself. Answer about the content itself. Ignore what the camera-person is doing.
        \begin{itemize}[leftmargin=*]
            \item E.g., "There are two people, i.e. a running child, and the cameraman." / "The camera is moving fast." are not good answers.
        \end{itemize}
        \item When you enter numbers, please enter digits instead of text.
        \begin{itemize}[leftmargin=*]
            \item "Seventeen" -> "17"
        \end{itemize}
        \item Use your best judgment.
    \end{itemize}
    \item How to provide video evidence
    \begin{itemize}[leftmargin=*]
        \item The video evidence consists of all the frame intervals of the video that support the answer to your given question.
        \item You need to provide at least one video evidence clip (interval within the video) for each question.
        \item You need to provide both the start point and end point for all the video evidence you identify in the video;
        \item You can use your mouse or ←/→ key to click or drag the process bars of the start point and end point. When you click or drag the bar, the above video will change accordingly, so you could locate the points according to the video screen.
        \item For each video evidence clip, the end point should be greater than zero, and the end point should be greater or equal to the start point.
        \item The video evidence clips (the time gap between the start point and the end point) should only cover the actual evidence and not more (in other words, it should be as short as possible).
    \end{itemize}
\end{enumerate}

\subsection{Annotation Interface}
\label{appendix-subsec:annotation-interface}

\begin{figure}
\centering
\includegraphics{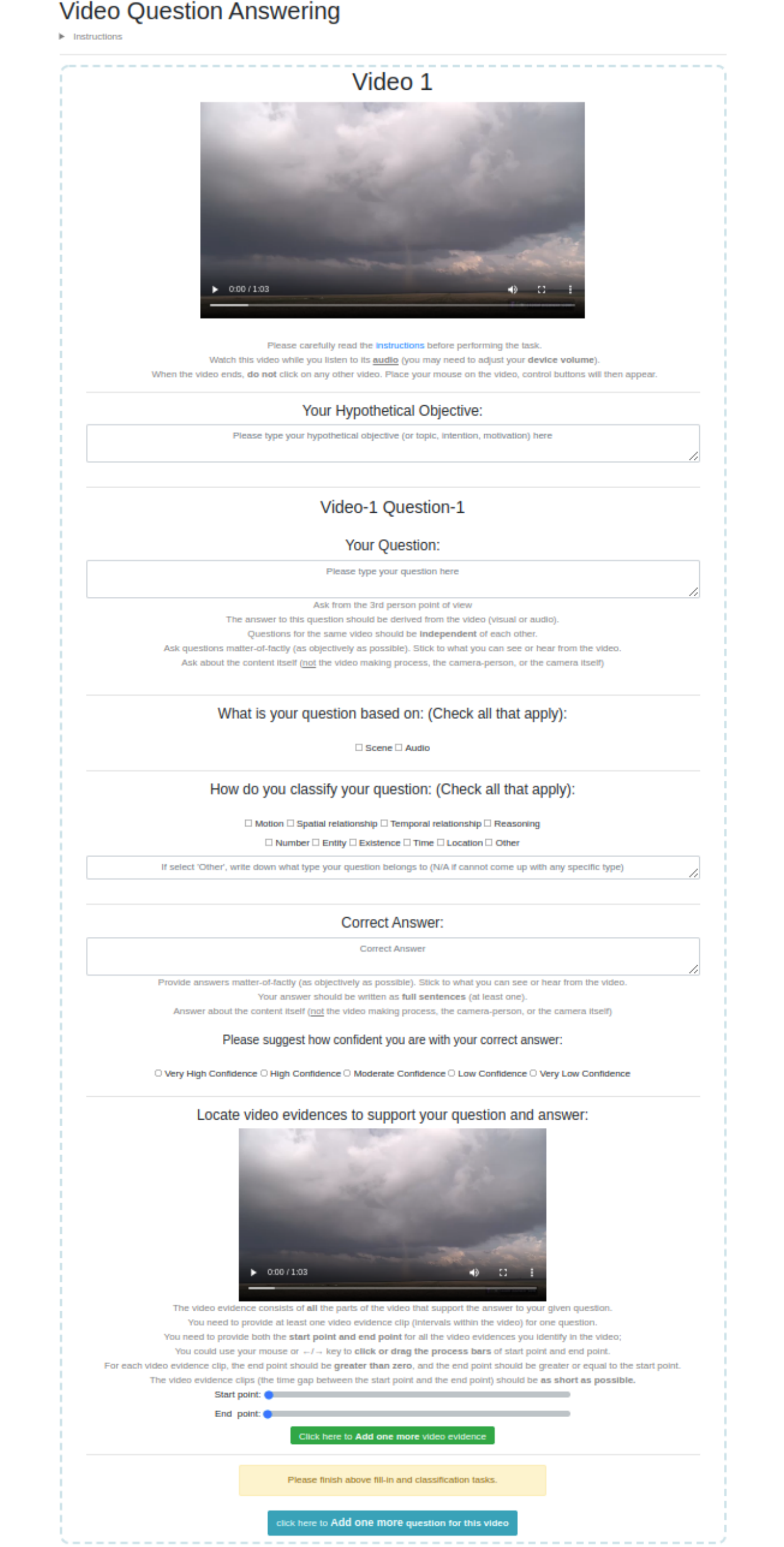}
\caption{Interface for Phase 1 annotation. After watching the video, annotators provide a \textbf{motivation}, ask \textbf{questions} and provide corresponding \textbf{answers} by filling the blank. They provide parts of the videos as \textbf{evidence} to support each of the question-answer pairs by dragging the moving bar.}
\label{fig:phase-1-interface}
\end{figure}

\begin{figure}
\centering
\includegraphics{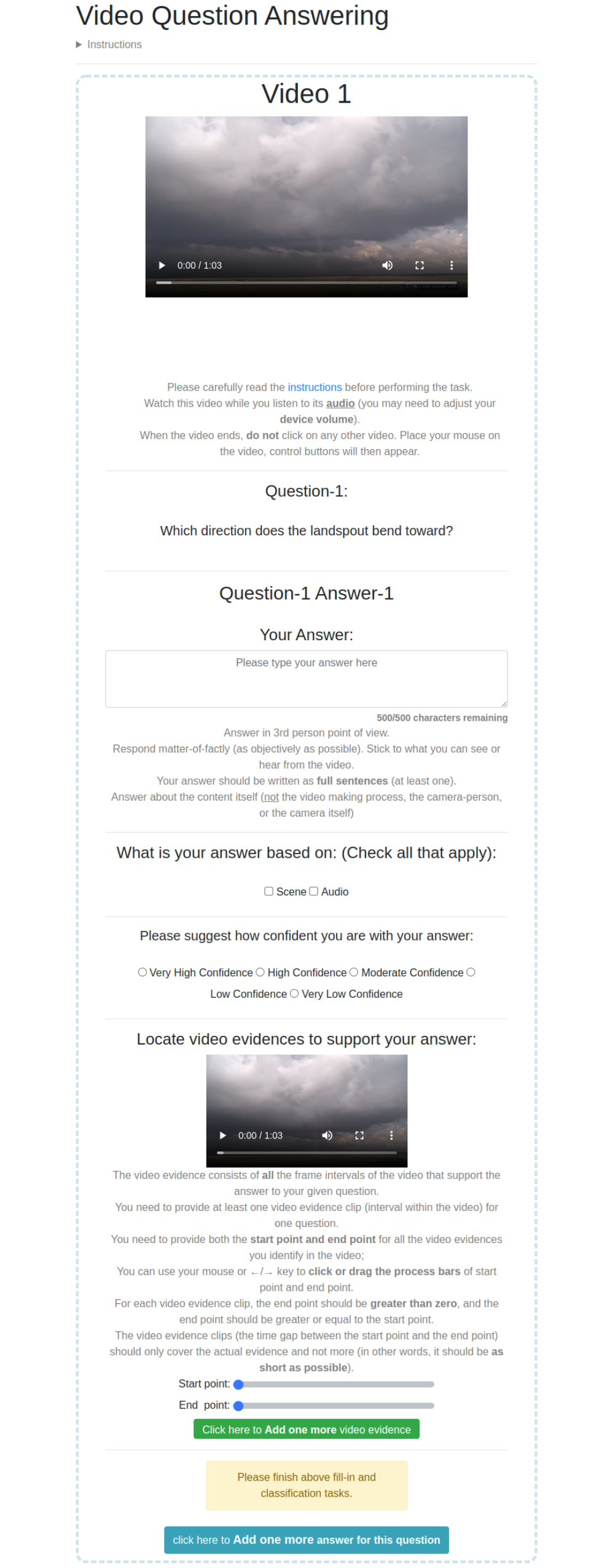}
\caption{Interface for Phase 2 annotation. After watching the video and given the question from Phase 1, annotators provide \textbf{answers} with the corresponding \textbf{evidence}.}
\label{fig:phase-2-interface}
\end{figure}

\Cref{fig:phase-1-interface} shows the annotation interface for Phase 1. \Cref{fig:phase-2-interface} shows the annotation interface for Phase 2. 

\subsection{Pilot Study Comparison between Annotations from Experts v.s. Non-Expert}
\label{appendix-subsec:pilot-study}

Before the formal annotation, we compare the non-experts and experts’ annotations for both phases. For Phase 1, we randomly selected 45 videos from each domain to be annotated by both the experts and crowdworkers. Following~\citet{castro2021fill}, we set the AWS annotation qualification as HIT approve rate \textgreater 92\%, the number of HITs approved \textgreater 1000, the location is either Canada or U.S., and the reward as \$6/HIT (around \$9/h).

After annotation, two authors of this paper who do not know the source of annotation evaluate and score in terms of Relevance, Interestingness, and Professionality for each annotation from 0 to 3. We define Relevance, Interestingness, and Professionality as follows:

\begin{itemize}[leftmargin=*]
    \item \textbf{Relevance}: how relevant a question and an answer are to the video. Good relevance indicates that the question is related to the video and focuses on the major events, objects, or people in the video. A relevant answer should address the question and can be derived from this video.
    \item \textbf{Interestingness}: whether the question interests you. In other words, whether you are interested in the question and answer, given a video.
    
    \item \textbf{Professionality}: how detailed and precise the question and answer are. Good professionality can be demonstrated by the precise usage of terminologies and numbers, and accurate description in the answer.
    
    \item \textbf{Overall Score}: the average score of the score for Relevance, Interestingness, and Professionality.
\end{itemize}

For each category, the higher score indicates the better the annotation demonstrates that characteristic. \cref{tab:question-pilot} lists the scores and \cref{tab:question-pilot-case} presents some annotation examples. From both the empirical and numerical results, we could see there is a significant quality gap for the annotation from experts versus from crowdworkers. Therefore, we decide to employ domain experts for Phase 1. 

\begin{table}
\begin{tabular}{lrrrr}
\toprule
       & \multicolumn{1}{l}{R} & \multicolumn{1}{l}{I} & \multicolumn{1}{l}{P} & \multicolumn{1}{l}{Overall} \\
       \midrule
expert & 2.7                  & 2.5                  & 2.1                  & 2.4                   \\
crowd  & 0.8                 & 0.7                 & 0.5                 & 0.7                  \\
\bottomrule
\end{tabular}
\caption{Average scores of the pilot study for Phase 1 (from 0 to 3). \textbf{R}: Relevance; \textbf{I}: Interestingness; \textbf{P}: Professionality; \textbf{Overall}:Overall Score}
\label{tab:question-pilot}
\end{table}

\begin{table*}
\small
\begin{tabular}{llll}
\toprule
       & Objective                         & Question                                   & Answer                \\
\midrule
E & Precipitation                     & What types of precipitation are occurring? & Rain and hail.        \\
C  & Very like                         & Nice                                       & Nice                  \\
\midrule
E & I want to learn about the people  & What type of weapons are they carrying?    & M4's                  \\
C  & The soldiers are caught on the ship. & What they are doing in this video?         & They caught the ship. \\
\midrule
E & Storm                             & Where is the storm?                        & In a field.           \\
C  & Motivation                       & 5                      & Very amazing         \\
\bottomrule
\end{tabular}
\caption{Examples in pilot study for Phase 1. \textbf{E}: Expert; \textbf{C}: Crowd}
\label{tab:question-pilot-case}
\end{table*}




 

\begin{table}
\small
\begin{tabular}{lrrrrrrr}
\toprule
Annotator & R1    & R2   & RL   & IOU-F1\\
\midrule
Expert   & 23.63 & 8.05 & 21.22 & 12.24\\
Crowd    & 20.03 & 3.24 & 17.69 & 8.50 \\
\bottomrule
\end{tabular}
\caption{ROUGE and IOU-F1 scores for the pilot study in Phase 2. Note that the scores here are lower than the scores for the human baselines in \cref{tab:few-shot-qa-results-10-epochs,tab:few-shot-evidence-results-10-epochs}. This is because we only compare the collected answers to a single answer here, while in \cref{tab:few-shot-qa-results-10-epochs,tab:few-shot-evidence-results-10-epochs} we calculate the average scores of one annotator against the remaining as described in \cref{sec:video-question-answering}.}
\label{tab:pilot-score}
\end{table}

For Phase 2, we randomly select 104 \dn{Geography} videos and questions from the questions annotated in Phase 1 to be annotated by both experts and crowdworkers. Moreover, we set the reward as \$3/HIT(around \$9/h) and employ the AWS \textbf{Master}\footnote{\url{https://www.mturk.com/worker/help\#what_is_master_worker}} as the crowdworkers. \Cref{tab:pilot-score} lists the results of the pilot study for Phase 2. According to \cref{tab:pilot-score}, crowdworkers perform similarly to experts in Phase 2. Considering the annotation efficiency, we decide to employ both experts and crowdworkers to annotate more diversified answers for each question in Phase 2. Note that the ROUGE scores in \cref{tab:pilot-score} are lower than the scores for the human baselines in \cref{tab:few-shot-qa-results-10-epochs,tab:few-shot-evidence-results-10-epochs}. This is because we only compare the collected answers to a single answer in \cref{tab:pilot-score}, while in \cref{tab:few-shot-qa-results-10-epochs,tab:few-shot-evidence-results-10-epochs}, we calculate the average scores of one annotator against the remaining as described in \cref{sec:video-question-answering}.

\subsection{Question and Answer Correction}
\label{appendix-subsec:question-answer-corrections}
After we collect annotation from Phase 1, the authors of this paper check the quality of the collected question and answers and modify the question and answers accordingly. Specifically, we:
\begin{itemize}[leftmargin=*]
    \item Delete the questions that can be answered without watching the video (e.g. \texttt{Q: ``If water can get through the hut's roof; can the wind go through the hut's roof?''}, \texttt{A: ``Yes the wind can go through the hut's roof.''})
    
    \item Modify the question or the answer to 3rd person view (e.g. change \texttt{Q: ``Do we have aircraft that we can do a touch and go landing like a helicopter?''} to \texttt{Q: ``Do they have aircraft that can do a touch and go landing like a helicopter.''})
    
    \item Exclude the man holding the camera in the answer if it is a first-person view video.
    
    \item Modify questions that are not independently asked (e.g.\texttt{``Where are they?''}, where ``they'' refers to the ``paved and unpaved roads'' in the previous question. Therefore, we change the question to \texttt{``Where are the roads?''} )
    
    \item Split questions that include multiple sub-questions into several questions.
    
\end{itemize}

Some of the annotators from Phase 2 do not annotate any evidence (leaving the evidence from the start to the end of the video). Thus, we empirically filter out evidence longer than $1/4$ of the video.

\subsection{Annotator Information}
\label{appendix-subsec:annotator-information}

\Cref{tab:annotator_expertise} shows the expertise of each expert, together with their assigned domains of annotation and the number of questions they annotate in their assigned domains in Phase 1.

\begin{table}
\small
\begin{tabular}{ccL{3.2cm}}
\toprule
Annotator ID & Expertise         & Assigned Domains (\# Q)\\
\midrule
0 & Geography & Geography (94) ; Natural Disaster (187) \\

\rg 1 & Geography & Geography (16) ; Human Survival (74) \\

2 & Veteran & Military (26) ; Human Survival (146) \\

\rg 3 & Veteran & Military (70) ; Human Survival (89) \\

4 & Veteran & Military (12) \\

\rg 5 & Veteran & Military (8) \\

6 & Veteran & Military (85) \\

\rg 7 & Biology & Agriculture (88) \\

8 & Biology & Agriculture (21) \\
\bottomrule
\end{tabular}
\caption{Information for expert annotators who annotate the questions, together with their assigned domains and number of questions (\# Q) in the parentheses.}
\label{tab:annotator_expertise}
\end{table}

\subsection{Dataset Analysis}
\label{appendix-subsec:dataset-analysis}

\begin{figure*}[t]
\centering
\includegraphics{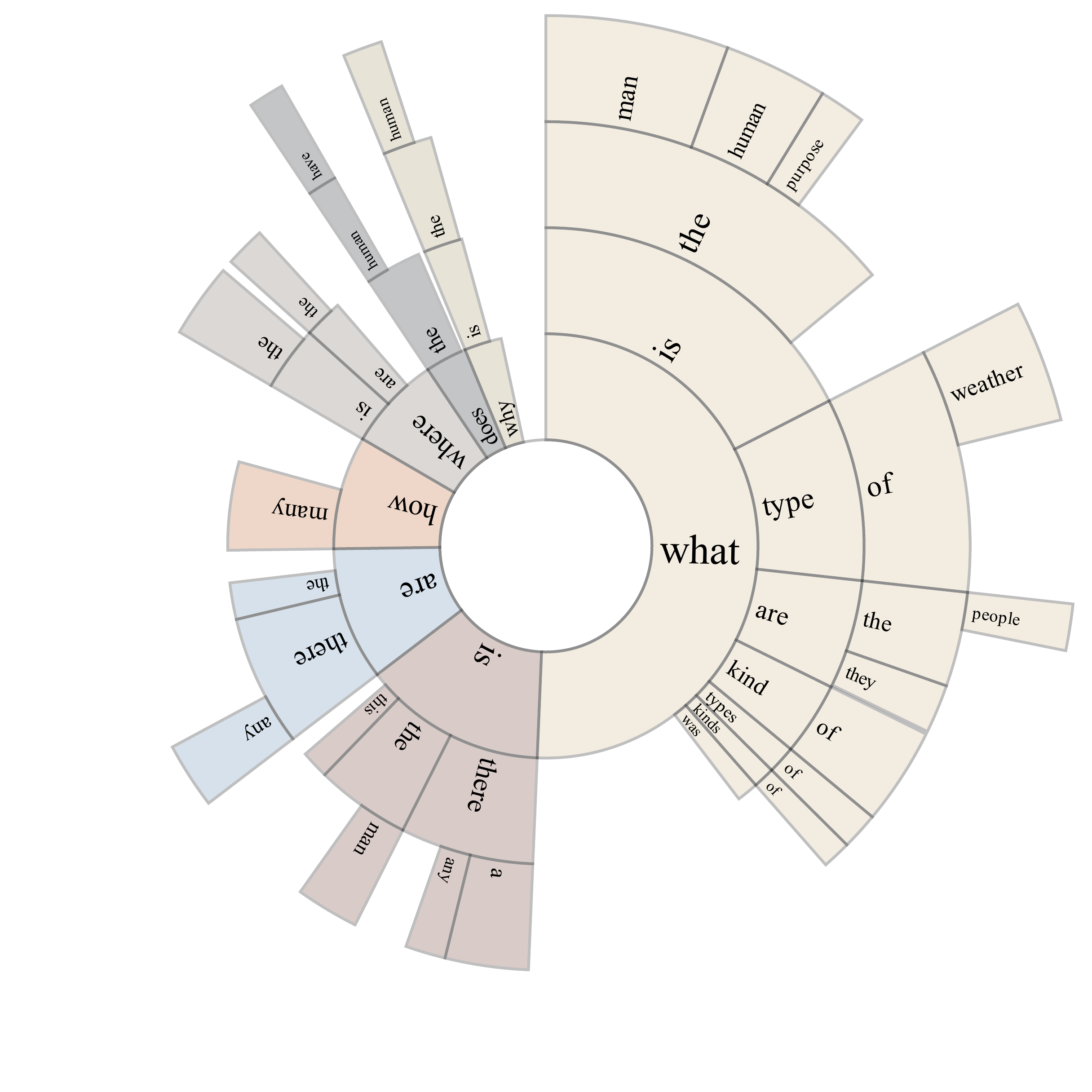}
\caption{Distribution of questions by the first four tokens. The ordering of words starts from the center to outside.}
\label{fig:question-sunburst}
\end{figure*}

\begin{table}
\small
\begin{tabular}{llll}
\toprule
Domain           & top1     & top2         & top3       \\
\midrule
\dn{Agriculture}      & farm     & agricultural & understand \\
\dn{Natural Disaster} & weather  & people       & flooding   \\
\dn{Human Survival}   & man      & determine    & human      \\
\dn{Geography}        & people   & topography   & water      \\
\dn{Military}         & military & aircraft     & determine \\
\bottomrule
\end{tabular}
\caption{Most common 3 words for each domain after removing stop-words.}
\label{tab:objective-most-tokens}
\end{table}

\Cref{fig:question-sunburst} presents question distributions in terms of words.

\paragraph{Questions Types.} \cref{tab:objective-most-tokens} examines the frequent words for each domain, which demonstrates the characteristics of the domain. Take \dn{Natural Disaster} as an example, the 3 most frequent words are used in 20.63\% of sentences. Besides, \cref{fig:question-type} in \cref{sec:wildqa-dataset} lists the annotators’ self-reported question types. One thing we observe is that questions that start with ``What'' possess a large proportion of all the questions. Such questions might be hard to classify into certain question typs~\cite{castro2020lifeqa}, so we allow annotators to choose multiple question types for a single question. Empirically speaking, questions that start with ``is(are)''/``where''/``how many'' are commonly relevant to "Existence"/"Location"/"Number" questions. In our dataset, their distribution trend (``is(are)'': 24.13\% > ``where'': 7.21\% > ``how many'': 4.48\%) is akin to the trend of the distribution of the reported question types (``Existence'': 45.20\% > ``Location'': 12.23\% > ``Number'': 4.59\%). Moreover, although we have ``human'', ``man'' and ``people'' as the most frequent words in some domains, the most frequent words in domains such as \dn{Military} are ``military'', and ``aircraft'', which demonstrates that our dataset does not only focus on human interactions as most of the existing datasets do.

\begin{figure}
\centering
\includegraphics{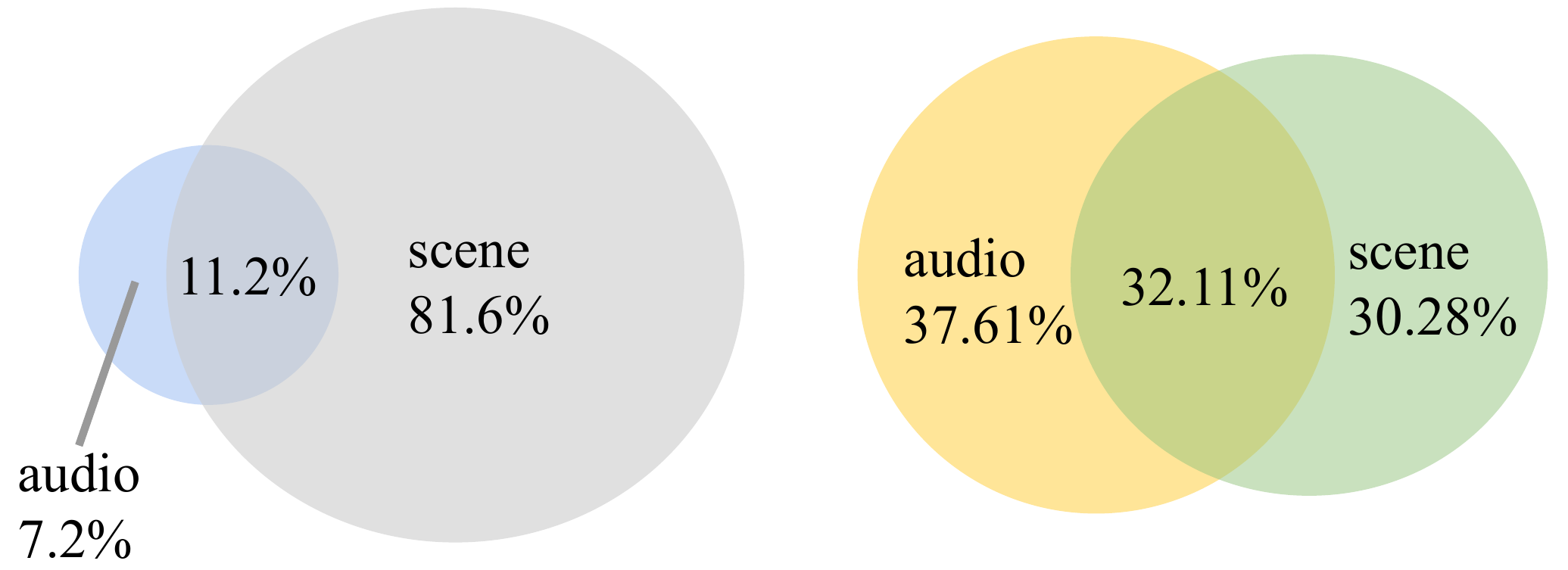}
\caption{Venn diagrams showing whether the question depends on visual (scene) or audio from the original video. The left is for the entire dataset, while the right is for the \dn{Agriculture} domain.}
\label{fig:qa-base-on}
\end{figure}

\paragraph{Information Needed.} As shown in the left Venn figure in \cref{fig:qa-base-on}, generally, most questions are based on the visual (scene). 
Such a distribution is also justified by the distribution of the question types. The dominant question types we have in \cref{fig:question-type} are \textit{Motion}, \textit{Spatial}, \textit{Existence} and \textit{Entity}, which typically focus on visual information. However, in \dn{Agriculture} (the right Venn figure in \cref{fig:qa-base-on}), the audio-based questions take more portion, because videos in \dn{Agriculture} usually focus on farming tips, instructions for using tools, etc. In this paper, we do not experiment with models that use audio or transcripts from the video. Future research might look into letting models use audio and transcripts on our dataset.

\paragraph{Answer Similarity/Diversity.} We have similar and diversified answers collected in our dataset. \Cref{fig:more-dataset-examples} gives 2 examples: answers from the upper example are similar to each other; for the lower example, answers diverse a lot between Phase 1 and Phase 2 annotations or even within Phase 2. However, all of the answers are acceptable given the video. The similarity demonstrates the reliability of the Phase 2 annotation. Meanwhile, the diversified answers help to better evaluate models.

\section{Annotation Statistics}

\Cref{tab:1st-anno-info,tab:2ed-anno-info} list the statistics for annotation in Phase 1 and Phase 2, respectively.

\begin{table}
\begin{tabular}{lc}
\toprule
Videos              & 369                                 \\
Duration (s)        & 71.22 $\pm$ 26.47 \\
\midrule
Questions           & 916                                 \\
Question per video  & 2.48 $\pm$ 1.38                          \\
Question length (\#tokens)    & 7.09 $\pm$ 2.60                          \\
Answer length (\#tokens)      & 8.62 $\pm$ 8.90                          \\
\midrule
Evidence per answer & 1.53 $\pm$ 0.76                          \\
Evidence length (s)    & 9.09 $\pm$ 13.45                        \\
\bottomrule
\end{tabular}
\caption{Annotation statistics for Phase 1. ``\#tokens'' represent the number of tokens.}
\label{tab:1st-anno-info}
\end{table}

\begin{table}
\begin{tabular}{lc}
\toprule
Crowd annotated answers & 932  \\
Expert annotated answers & 182  \\
Total    & 1114              \\
\midrule
Answer per question & 1.22 $\pm$ 0.69                  \\
Answer length (\#tokens)      & 9.45 $\pm$ 7.46                 \\
\midrule
Evidence per answer & 0.89 $\pm$ 0.72                  \\
Evidence length (s)    & 10.43 $\pm$ 5.81                \\
\bottomrule
\end{tabular}
\caption{Annotation statistics for Phase 2. ``\#okens'' represents the number of tokens.}
\label{tab:2ed-anno-info}
\end{table}

\section{Details of Multi-task Learning}
\label{sec:details-of-multi-task-learning}

\begin{table*}
\small
\begin{tabular}{lrrrr}
 \toprule
 $\beta$  & R\@1 & R\@2 & R\@L & IOU-F1 \\
 \midrule
 0.5 &\textbf{33.8} $\pm$ \textbf{0.8}&\textbf{18.5} $\pm$ \textbf{0.7}&\textbf{32.5} $\pm$ \textbf{0.8} &\textbf{3.7} $\pm$ \textbf{2.4} \\
 
 1.0 &32.2 $\pm$ 0.7&17.6 $\pm$ 0.5&31.0 $\pm$ 0.6 &1.9 $\pm$ 1.7\\
 
 1.5 &33.8 $\pm$ 0.3&18.0 $\pm$ 0.9&32.5 $\pm$ 0.3 &1.5 $\pm$ 0.1\\
 
 \bottomrule
\end{tabular}
\caption{We set $\alpha=1$ throughout all the experiments, and report the corresponding \mn{Multi}$_\text{T+V,SE}$ performances on Video QA (ROUGE scores) and Video Evidence Selection (IOU-F1 scores). We highlight the row we report in \cref{tab:few-shot-qa-results-10-epochs} in \cref{subsec:qa-results} and \cref{tab:few-shot-qa-results-10-epochs} in \cref{subsec:qa-results}.}
\label{tab:multi-task-parameter-selection-for-evidence-se}
\end{table*}

\begin{table*}
\small
\begin{tabular}{lrrrr}
 \toprule
 $\beta$  & R\@1 & R\@2 & R\@L & IOU-F1 \\
 \midrule
 0.5 &\textbf{34.0} $\pm$ \textbf{0.5}&\textbf{18.8} $\pm$ \textbf{0.7}&\textbf{32.8} $\pm$ \textbf{0.6} &\textbf{1.2} $\pm$ \textbf{0.1}\\
 
  1.0 &33.4 $\pm$ 0.6&18.4 $\pm$ 0.2&32.1 $\pm$ 0.6&1.4 $\pm$ 0.3\\
 
 1.5 &32.8 $\pm$ 0.3&18.3 $\pm$ 0.3&31.7 $\pm$ 0.2&1.0 $\pm$ 0.2\\
 
 \bottomrule
\end{tabular}
\caption{We set $\alpha=1$ throughout all the experiments, and report the corresponding \mn{Multi}$_\text{T+V,IO}$ performances on Video QA (ROUGE scores) and Video Evidence Selection (IOU-F1 scores). We highlight the row we report in \cref{tab:few-shot-evidence-results-10-epochs} in \cref{subsec:evidence-results} and \cref{tab:few-shot-evidence-results-10-epochs} in \cref{subsec:evidence-results}.}
\label{tab:multi-task-parameter-selection-for-evidence-io}
\end{table*}

\Cref{tab:multi-task-parameter-selection-for-evidence-se,tab:multi-task-parameter-selection-for-evidence-io} report the model performances under different sets of $\alpha, \beta$ for \cref{eq:multi-task-learning}. We highlight the rows we report in \cref{tab:few-shot-qa-results-10-epochs} in \cref{subsec:qa-results}, \cref{tab:few-shot-qa-results-10-epochs} in \cref{subsec:qa-results}, \cref{tab:few-shot-evidence-results-10-epochs} in \cref{subsec:evidence-results}, and \cref{tab:few-shot-evidence-results-10-epochs} in \cref{subsec:evidence-results}.

\section{Experiment Results}
\label{appendix:experiment-results}

\begin{figure}
\centering
\includegraphics{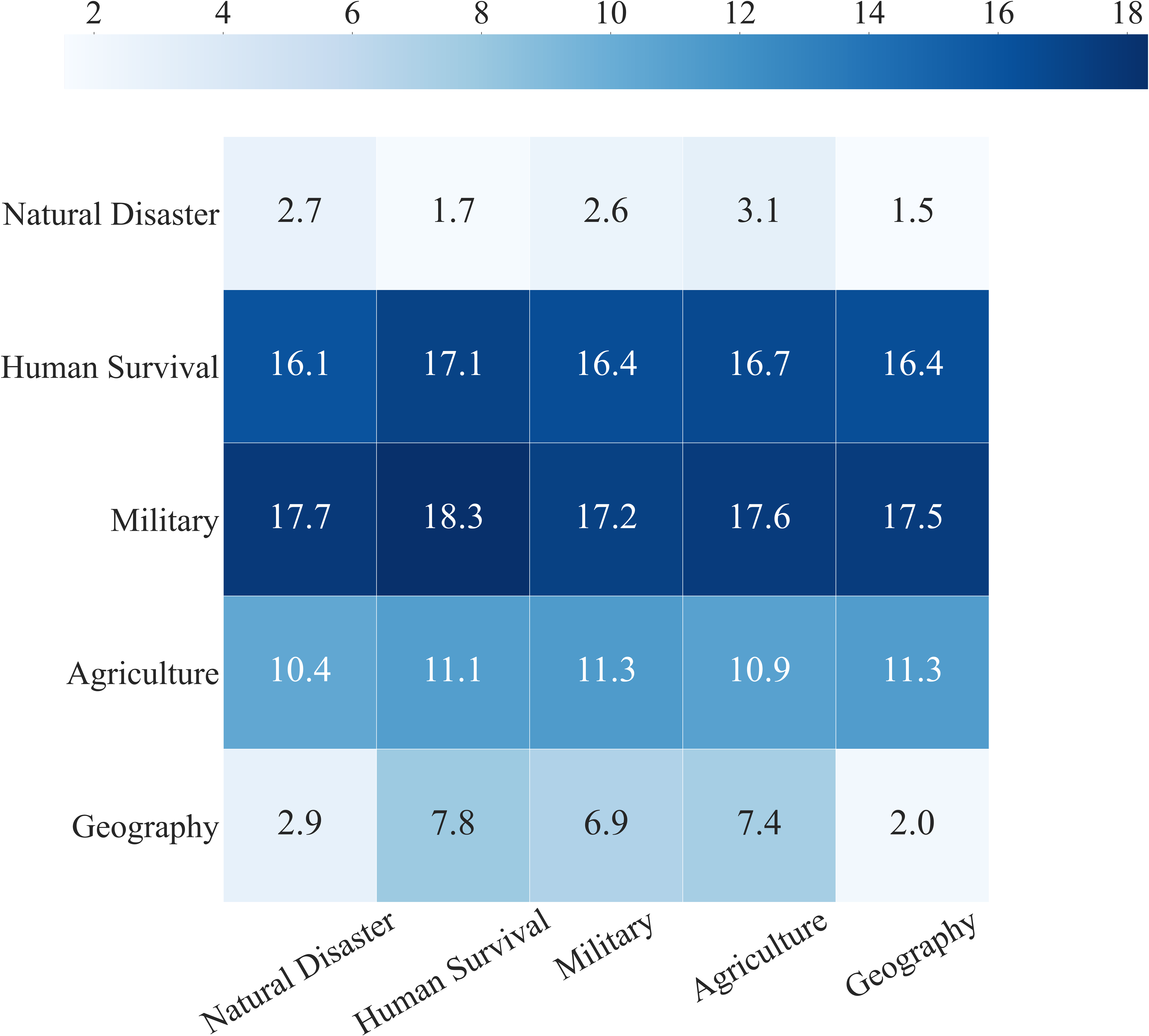}
\caption{\mn{Multi-Task} ROUGE-2 scores for Video QA when tuned on a single domain (y-axis) and tested against each domain (x-axis).}
\label{fig:rouge2-heatmap}
\end{figure}

\begin{figure}
\centering
\includegraphics{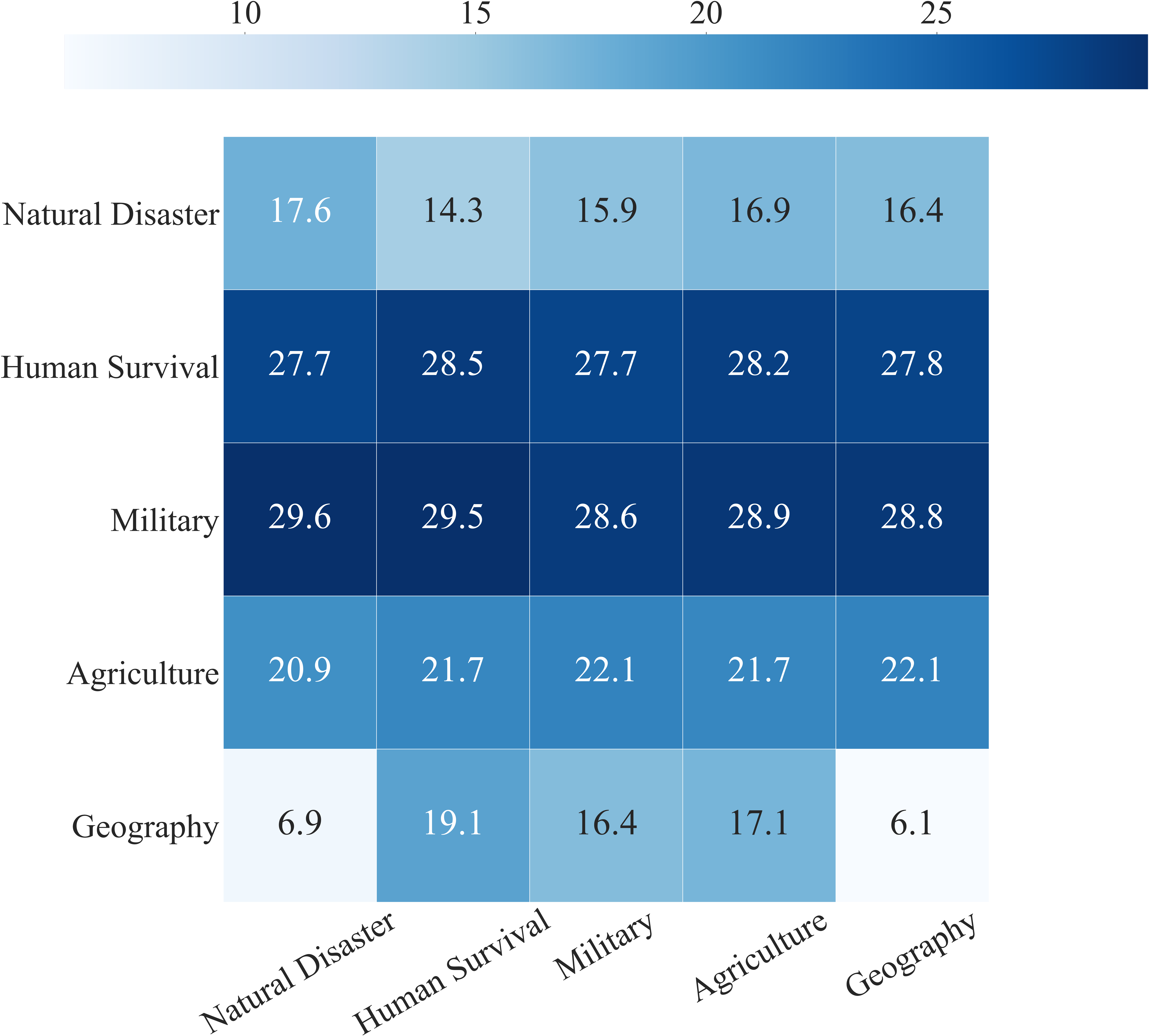}
\caption{\mn{Multi-Task} ROUGE-L scores for Video QA when tuned on a single domain (y-axis) and tested against each domain (x-axis).}
\label{fig:rougel-heatmap}
\end{figure}

\begin{figure}
\centering
\includegraphics{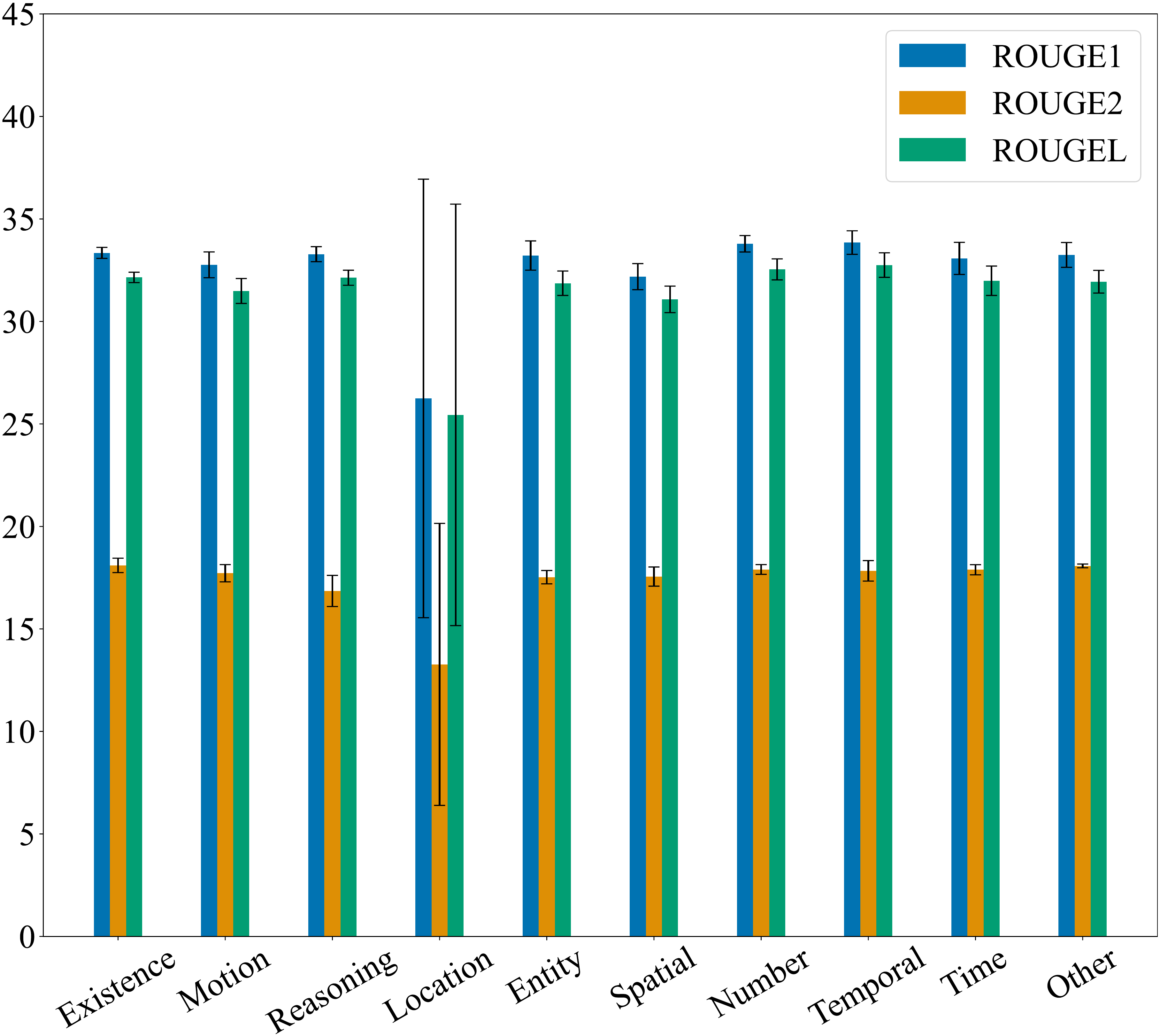}
\caption{\mn{Multi$_\text{T+V,SE}$} performance on different question types for Video QA. For each question type, we report ROUGE-1, ROUGE-2, and ROUGE-L scores from left to right. We can see that different ROUGE scores follow similar trends, we only report ROUGE-1 in \cref{tab:model-performance-type-performance-rouge-1-iou} in \cref{sec:analysis-and-discussion}.}
\label{fig:model-performance-type-performance}
\end{figure}

\Cref{fig:rouge2-heatmap,fig:rougel-heatmap} report \mn{Multi-Task} model's performance on Video QA by ROUGE-2, and ROUGE-L, respectively. \Cref{fig:model-performance-type-performance} demonstrates that ROUGE scores follow a similar trend as mentioned in \cref{sec:analysis-and-discussion}.

\subsection{Ablation Study on Video Evidence Selection}
\label{subsec: ablation-study-on-video-evidence-selection}

To investigate whether the vision part is indeed needed by the baseline models for the Video Evidence Selection task, we conduct an ablation study using \mn{T5 IO} and \mn{T5 SE} (introduced in \cref{sec:video-evidence-selection}). We take a random sequence of the same length as the original video sequence and feed the random sequence instead of the original video sequence to the model. \Cref{tab:random-comparison-evidence-results} shows the results of the comparison between these different settings. 
\mn{T5 IO} performs roughly the same as \mn{T5 IO$_\text{random}$}, which indicates that the model struggles to utilize visual information. \mn{T5 IO} even underperforms the random baseline which can achieve an IOU-F1 score of 2.5 $\pm$ 0.3 (as shown in Table~{tab:few-shot-evidence-results-10-epochs}). However, \mn{T5 SE} outperforms \mn{T5 SE$_\text{random}$}, suggesting that \mn{T5 SE} uses visual features to locate the evidence of the question. 

\begin{table}
\small
\begin{tabular}{lr}
 \toprule
 Model name & IOU-F1 \\
 \midrule

\mn{T5 IO$_\text{random}$} &1.1 $\pm$ 0.3\\

\mn{T5 IO}  &1.1 $\pm$ 0.2 \\

\mn{T5 SE$_\text{random}$} &2.7 $\pm$ 1.9\\

\mn{T5 SE} & \textbf{4.5} $\pm$ \textbf{0.8} \\

\bottomrule
\end{tabular}
\caption{Ablation study on Video Evidence Selection. We feed \mn{T5 IO$_\text{random}$} and \mn{T5 SE$_\text{random}$} the question concatenated with a random sequence, while we feed \mn{T5 IO} and \mn{T5 SE} the question with the actual video sequence.}
\label{tab:random-comparison-evidence-results}
\end{table}

\end{document}